\documentclass[12pt]{article}
\usepackage[round,sort,authoryear]{natbib}

\usepackage{amsmath,amssymb,amsthm,amsfonts, fullpage, setspace}

\usepackage{eucal}
\usepackage{subeqnarray,bm}
 \usepackage[utf8]{inputenc}
\usepackage{graphicx}[final]
\usepackage{psfrag} 
\usepackage{color,float}
\usepackage{pdfpages}

\usepackage{epstopdf}

\usepackage[colorlinks]{hyperref}

\usepackage{hyperref}
\hypersetup{colorlinks=red,citecolor=blue,pdfstartview=FitH, pdfpagemode=None}
\usepackage{multirow}
\usepackage[bottom]{footmisc}
\usepackage{ifthen} 
\usepackage{sectsty}
\usepackage{subfig}
\usepackage{float}
\usepackage[english]{babel}
\usepackage[nottoc]{tocbibind}
\usepackage{calrsfs}
\usepackage{soul}
 \DeclareGraphicsExtensions{.eps,.png,.pdf}
\usepackage{mwe}
\makeatletter
\def\maxwidth{ %
  \ifdim\Gin@nat@width>\linewidth
    \linewidth
  \else
    \Gin@nat@width
  \fi
}
\makeatother

\definecolor{fgcolor}{rgb}{0.345, 0.345, 0.345}

\allsectionsfont{\normalsize\bfseries}

\usepackage{framed}
\makeatletter
 {\par\unskip\endMakeFramed%
 \at@end@of@kframe}
\makeatother

\definecolor{shadecolor}{rgb}{.97, .97, .97}
\definecolor{messagecolor}{rgb}{0, 0, 0}
\definecolor{warningcolor}{rgb}{1, 0, 1}
\definecolor{errorcolor}{rgb}{1, 0, 0}

\usepackage{amsmath, amssymb, geometry, booktabs, graphicx}
\newtheorem{thm}{Theorem}[section]

\newtheorem{remark}[thm]{Remark}

\newcommand{\ds}{\displaystyle}

\newcommand{\abs}[1]{\left\vert#1\right\vert}
\newcommand{\set}[1]{\left\{#1\right\}}

\newcommand{\R}{\mathbb{R}}

\newcommand {\xs}{x^*}
\newcommand {\ys}{y^*}

\numberwithin{equation}{section}

\usepackage{calrsfs}

\allsectionsfont{\normalsize\bfseries}

\numberwithin{equation}{section}
\allowdisplaybreaks

\usepackage[table, dvipsnames]{xcolor}
\DeclareMathAlphabet{\pazocal}{OMS}{zplm}{m}{n}
    \newcounter{example}[section]

\title{Allee Synaptic Plasticity and Memory}
\author{Eddy Kwessi\\
Department of Mathematics\\
Trinity University, San Antonio, Texas}
\date{}					

\begin{document}
\maketitle
\begin{abstract}
Neural plasticity is fundamental to memory storage and retrieval in biological systems, yet existing models often fall short in addressing noise sensitivity and unbounded synaptic weight growth. This paper investigates the Allee-based nonlinear plasticity model, emphasizing its biologically inspired weight stabilization mechanisms, enhanced noise robustness, and critical thresholds for synaptic regulation. We analyze its performance in memory retention and pattern retrieval, demonstrating increased capacity and reliability compared to classical models like Hebbian and Oja’s rules. To address temporal limitations, we extend the model by integrating time-dependent dynamics, including eligibility traces and oscillatory inputs, resulting in improved retrieval accuracy and resilience in dynamic environments. This work bridges theoretical insights with practical implications, offering a robust framework for modeling neural adaptation and informing advances in artificial intelligence and neuroscience.
\end{abstract}

\section{Introduction}\label{sec1}
We recall that brain plasticity also known as neuroplasticity is the brain's ability to change and adapt throughout an individual's life. 
Neuroplasticity involves the reorganization of neural networks, changes in synaptic connections, and the creation of new neurons in specific brain regions. It is often categorized into two types. On the  one hand,  structural plasticity which refers to physical changes in the brain's structure, for instance growth or pruning of dendrites and axons (\cite{Riccomagno2015, Kirchner2024, Low2006}), changes in the sizes or density of gray matter regions \cite{Seminowicz2010, Henssen2019, Yankowitz2021}. On the other hand, functional plasticity refers to changes in the brain's functional activity, for instance the reallocation of functions from damaged areas to healthy areas after injury (\cite{Dancause2011, Nudo2013, Guggenmos2013}). Factors influencing brain plasticity include but are not limited to age (the brain is more plastic at an early age and less so with increasing age), experience learning (education, skill development), injury (stroke or trauma), environment (social interaction, stimulating environment), or biological factors (genetic predisposition, hormones, neuro-chemical changes). Brain plasticity occurs through a handful of mechanisms which include neurogenesis,  or the formation of new neurons (\cite{Eriksson1998, Kempermann2018, Tan2021}), rewiring or formation of new connections in response to learning or damage (\cite{Bennett2018, Radulescu2021}), compensatory mechanisms or the reallocation of tasks to different brain regions when a particular area is impaired (\cite{Kim2022, Lazzouni2014, Balbino2019}), or synaptic plasticity or the brain’s ability to strengthen or weaken synaptic connections over time, based on neural activity.  In this paper, we will focus on the latter. Modeling synaptic plasticity entails deriving mathematical equations that describe connections between neurons in response to activity.  This involves dynamically  linking  pre and post synapses via some weights in order to understand learning, memory, and adaptation in the brain. Many such models have been proposed in the literature and include, but are not limited to, the basic Hebbian, Oja (\cite{Oja1982}),  Bienestock-Cooper-Munro (\cite{Bienenstock1982}), Covariance  (\cite{Dayan2001}), spike-timing dependent plasticity (STDP) (\cite{Markram1997, Bi1998, AndradeTalavera2023}), homeostatic plasticity (\cite{Turrigiano2012}), short-term plasticity (\cite{Zucker2002}), biophysical models of plasticity (\cite{Abarbanel2003}), probabilistic models of plasticity (\cite{Tully2014}), reinforcement learning-based models (\cite{Biagiola2022}),  and Allee  models (\cite{Kwessi2022}).  The Allee model has a nonlinear weight regulation term that regulates extreme weight growth or decay, avoiding issues like unbounded growth seen in Hebbian models. It allows for the suppression of synaptic weights beyond a certain threshold, enabling more realistic stabilization of network dynamics. This is particularly useful for modeling systems with sharp transitions in activity or extinction, such as in  populations with an Allee effect in biology or neural systems that suppress weak synapses. By dynamically modulating synaptic weights, the Allee model can better differentiate between stored patterns, potentially increasing the memory capacity. The Allee model's nonlinear feedback can make it more robust to noise and perturbations, ensuring that synaptic weights remain within biologically plausible bounds. The threshold parameter provides more flexibility to adjust the balance between weight potentiation and depression, allowing the model to adapt to specific neural tasks or datasets. The model is well-suited for systems where neural plasticity exhibits bi-stability or multi-stability, mimicking biological systems that display discrete states of activation or depression. Weak weights are prevented from vanishing entirely, a limitation in Oja’s model; instead, they decay nonlinearly, reflecting biologically observed plasticity mechanisms that retain weak but functional connections. 

While Hebbian, Oja, and STDP models are grounded in experimental observations of synaptic plasticity, the Allee model has not yet been empirically validated in synaptic plasticity but is commonly applied in other biological systems like evolution (\cite{DAnniello2025}) and ecology (\cite{Dennis2015, Elaydi2018, Kwessi2023}). The Allee model, as presented, is a theoretical construct designed to explore the impact of nonlinear feedback and critical thresholds on memory stability.  It provides a flexible and analytically rich framework to test hypotheses about how neural systems might regulate weight growth, encode memory, and respond to perturbations using threshold-based dynamics analogous to those found in ecological systems.

 In this paper, we discuss a version of the Allee model that utilizes a  sigmoid gain function  rather than a linear one. The proposed  model   preserves its key property of nonlinear feedback. We also show that  incorporating temporal dependencies improves robustness to noise. Specifically, the key contributions of  this paper are: \\
1) Bifurcation and stability analyses reveal the presence of multiple dynamic regimes: stable fixed points, extinction, and oscillatory patterns (including Hopf bifurcations), unlike in linear models.\\
2) The model has  richer dynamics than traditional models with analytical derivations of fixed points, bifurcations, and memory-relevant attractors.\\
3) The model encodes memory via multiple attractors and stability conditions.\\
4) Time-dependent features like eligibility traces  further enhance retrieval fidelity and robustness under noise. These temporal dynamics elevate the accuracy of retrieval, approaching STDP-level performance.

The remainder of the paper is organized as follows: in Section \ref{ModelDescription}, we describe how the model is constructed.  In Section \ref{FixedPoints}, we discuss the dynamics, bifurcation analysis, pattern retrieval 
 of a single layer and single post-synaptic neuron system. In Section \ref{sec:4},  we discuss associative memory, temporal dynamics,  and noise robustness for a multiple-layers and  multiple post-synaptic neurons system. In Section \ref{sec:concl}, we will make our final comments. 
 
 \section{Materials and Methods} \label{ModelDescription}
 We will describe the model under consideration  and then explain the intuition behind its construction. 
\subsection{Model description} 
Let $L$ be a positive integer representing the number of layers,  $N_u$ and  $N_v$ be given positive integers representing respectively the number of pre-synaptic and post-synaptic neurons, respectively.  For sake of completeness and self-containment,  we recall that the Allee model proposed in  \cite{Kwessi2022}, is as follows:
\begin{equation}\label{eqn:Alleerule1}
\begin{cases}
\ds \tau_{_{\bf v}} \frac{d \bf v}{dt} \vspace{0.25cm}&=-{\bf v}+T({\bf W, M, u,v})\\
\ds \tau_{_{\bf W}} \frac{d {\bf W}}{dt} &={\bf v}^T\left({\bf u}-K^{-1}{\bf  W}{\bf v}\right)\left({\bf 1}-A ({\bf W}^T{\bf W})^{-1}\right)
\end{cases}\;,
\end{equation}
where {\bf u} represents an $L\times N_u$ matrix  of pre-synaptic neurons, {\bf v} and $L\times N_v$ matrix of post-synaptic neurons, {\bf W} an $L\times N_u \times L\times N_v$ block matrix of pre-synaptic weights,  {\bf M} an $L\times N_v \times L\times N_v$ block matrix of post-synaptic weights, $\tau_{_{\bf v}}$ and $\tau_{_{\bf W}}$ represent respectively the time scales of the firing-rate dynamics of $\bf v$ and ${\bf W}$, and  $T({\bf W, M, u,v})$ is known as a gain function. In the sequel, we will use the sigmoid gain function $T({\bf W, M, u,v})=G({\bf W}^T{\bf u}+{\bf W}^T{\bf v})$, where $G(x)=(1+e^{-x})^{-1}$. We note that  in   \cite{Kwessi2022},  a linear gain function $T({\bf W, M, u,v})={\bf W}^T{\bf u}+{\bf W}^T{\bf v}$ was considered. The justification for this choice is that  synaptic plasticity rates are much lower  that  linear rate, see for instance \cite{Dayan2001}, page 285.
It is noteworthy that  instead of the sigmoid function,  one could also consider the Soboleva gain function $G_s(x)=(e^{ax}-e^{-bx})(e^{cx}+e^{-dx})^{-1}$, which for adequately chosen parameters $a, b, c$ and $d$, could produce firing rates much lower than sigmoid firing rates. 
\begin{figure}[htbp] 
   \centering
   \includegraphics[width=4in]{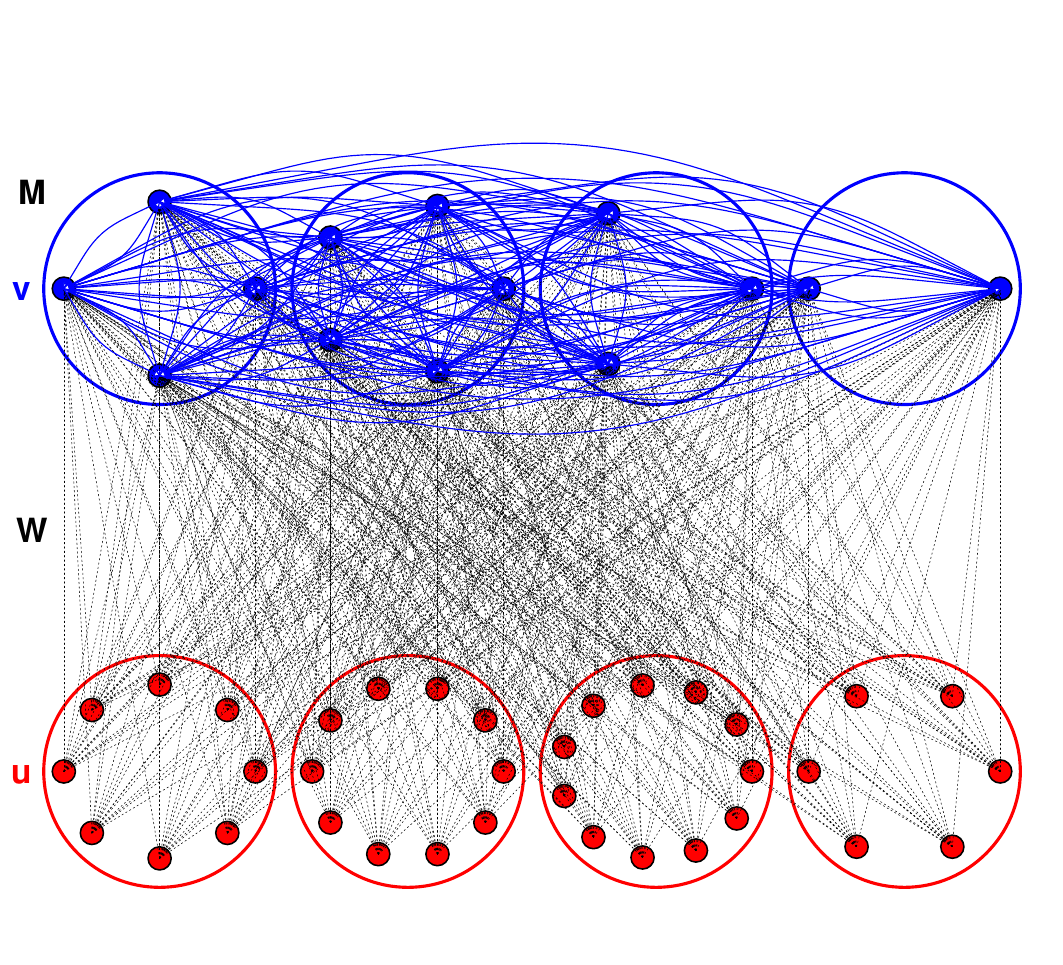} 
   \caption{Representation of the system with $L=3$ layers, with respective 8, 10,  and 11,  6 pre-synaptic neurons ${\bf u}$ ($N_u=35$), 3, 5, 3, and 2 post-synaptic neurons ${\bf v}$ ($N_v=13$), with weight  matrix $W$ (dashed lines), and post-synaptic connection matrix $M$ (blue curves).}
   \label{fig:system}
\end{figure}
\subsection{Intuition behind the model}

Suppose $L=1, N_v=1, N_u>1$,  and  ${\bf W}\neq {\bf 0}$. 
Since $N_v=1$, then ${\bf v}$ and ${\bf M}$  are scalars so let  $x=v$ be a post-synaptic neuron  with  a single recurrent connection weight  ${\bf M}=m$. We let  ${\bf u}=({\bf u}_1, {\bf u}_2, \cdots, {\bf u}_{N_u} )$ be a  vector of pre-synaptic neurons with feedforward weights ${\bf W}=(W_{i})_{1\leq i\leq N_u}$. We let  the scalar  $u=\|\bf u\|\cos(\theta)$, where $\theta$ is the angle between ${\bf W} $ and ${\bf u}$.  Multiplying the second equation in \eqref{eqn:Alleerule1} by $2{\bf W}^T$, we obtain
\[
\frac{d \|{\bf W}\|^2}{dt}=2{\bf W}^T{\bf v}^T ({\bf u}-K^{-1}{\bf Wv})(1-A\|{\bf W}\|^{-2})\;.
\]
Noting that ${\bf v}=v$, we obtain
\begin{eqnarray*}
\frac{d \|{\bf W}\|^2}{dt}&=&2v ({\bf W}^T{\bf u}-vK^{-1}{\bf W}^T{\bf W})(1-A\|{\bf W}\|^{-2})\\
&=&2v(\|{\bf W}\|\|{\bf u}\|\cos(\theta)-vK^{-1}\|{\bf W}\|^2)(1-A\|{\bf W}\|^{-2})\\
&=& 2v\left(u\|{\bf W}\|-\frac{v\|{\bf W}\|^2}{K}\right)\left(1-\frac{A}{\|{\bf W}\|^2}\right)\;.
\end{eqnarray*}
Rewriting the system  in terms of $x=v$ and $y=\|{\bf W}\|^2$,  and  assuming  for simplicity that $\tau_{\bf v}=1$ and $\tau_{\bf W}=2$, we obtain the equation 
\begin{equation}\label{eqn:AlleeModel1}
\begin{cases}
 \frac{d x}{dt} \vspace{0.25cm}&=f(x,y)=-x+G(u\sqrt{y}+mx)\\
 \frac{d y}{dt} &=g(x,y)=x\left(u\sqrt{y}-\frac{xy}{K}\right)\left(1-\frac{A}{y}\right)
\end{cases}\;.
\end{equation}
It is important to note that in the above equation, the parameter $K^{-1}$ represents  the interaction intensity  between $x$ and $y$ and $u$ and  plays the role of growth rate regulator of the dynamic of the length of weight $y$, with $y>0$.  What this shows is that equation \ref{eqn:Alleerule1} is just a block-matrix version of this model that accommodates multiple layers and interacting neurons.
\section{Results}  
We first revisit  simplest mathematical case of a single-layer and single post-synaptic neuron and highlight new findings regarding bifurcation analysis. Then we will discuss the impact of multiple neurons and multiple layers  on the robustness to noise. 
\subsection{Single layer and single post-synaptic neuron model}  \label{FixedPoints}
\subsubsection{Fixed Points and stability analysis}
The fixed points of the model \eqref{eqn:AlleeModel1} above are intersections of the isoclines $\frac{dx}{dt}=0,~ \frac{dy}{dt}=0$.
Since $G(v)>0$ for all $v\in \R$, setting  $x=0$ in $\frac{dy}{dt}=0$ would imply that $G(u\sqrt{y})=0$, which is impossible. Therefore, the have only nontrivial fixed points $(x_*,y_*)$ are given as $\left(x_1^*=G(\frac{u^2K}{x_1^*}+mx_1^*), y_1^*=\left(\frac{uK}{x_1^*}\right)^2\right)$ and $\left(x_2^*=G(u\sqrt{A}+mx_2^*),y_2^*=A\right)$. We remark that in case $A=\left(\frac{uK}{x_1^*}\right)^2$, the two fixed points are identical.
Since the system is two-dimensional and the origin is not a fixed point, the fixed points lie in the interior of the first quadrant. For a bounded  system, there are  three possibilities: one of the fixed point is stable (an attractor) and the other is unstable (a saddle)  and vice-versa, or there exist a  only one fixed point, which is globally asymptotically  stable if $A=0$ and unstable if $A>0$.  Note that in the latter case,  there would be a saddle-node bifurcation occurring after the collision of the two fixed points. To formalize the discussion, we state the following result :

 \begin{thm}\label{thm1} Suppose that  $A, K, u, m$ are absolute constants. 
 
Put
 \begin{eqnarray*}
\ds \tau_A&=&\frac{uK}{\sqrt{A}}\;, \\
x_A &=&G(m\sqrt{A}+mx_A)\;.
 \end{eqnarray*}
  Then the interior fixed point  $(x_A,A)$ is locally asymptotically stable  and  $(x_1^*,y_1^*)$ is a unstable if 
 \begin{itemize}
 \item if    $m<4$ and $x_A>\tau_A$, or 
 \item if  $m>4$  $m>4$, and $x_A\in (0, v_{A1})\cup (v_{A2},1)$, for some $0<v_{A1}<v_{A2}<1$.
 \end{itemize}
 \end{thm}

\begin{figure}[H]
 \centering \resizebox{1\textwidth}{!}{\begin{minipage}{1.3\textwidth}
 \begin{tabular}{cc}
 (a) & (b)\\
 \includegraphics[scale=0.45]{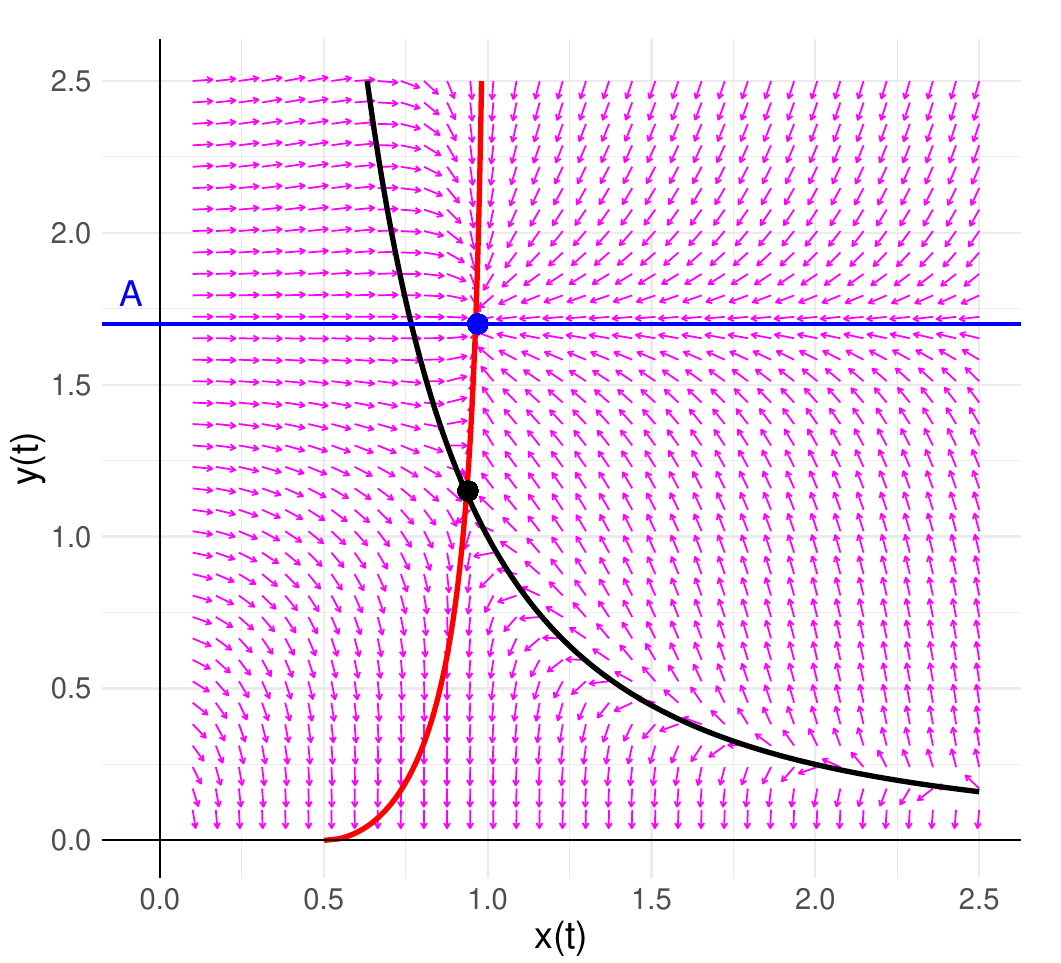} &\includegraphics[scale=0.45]{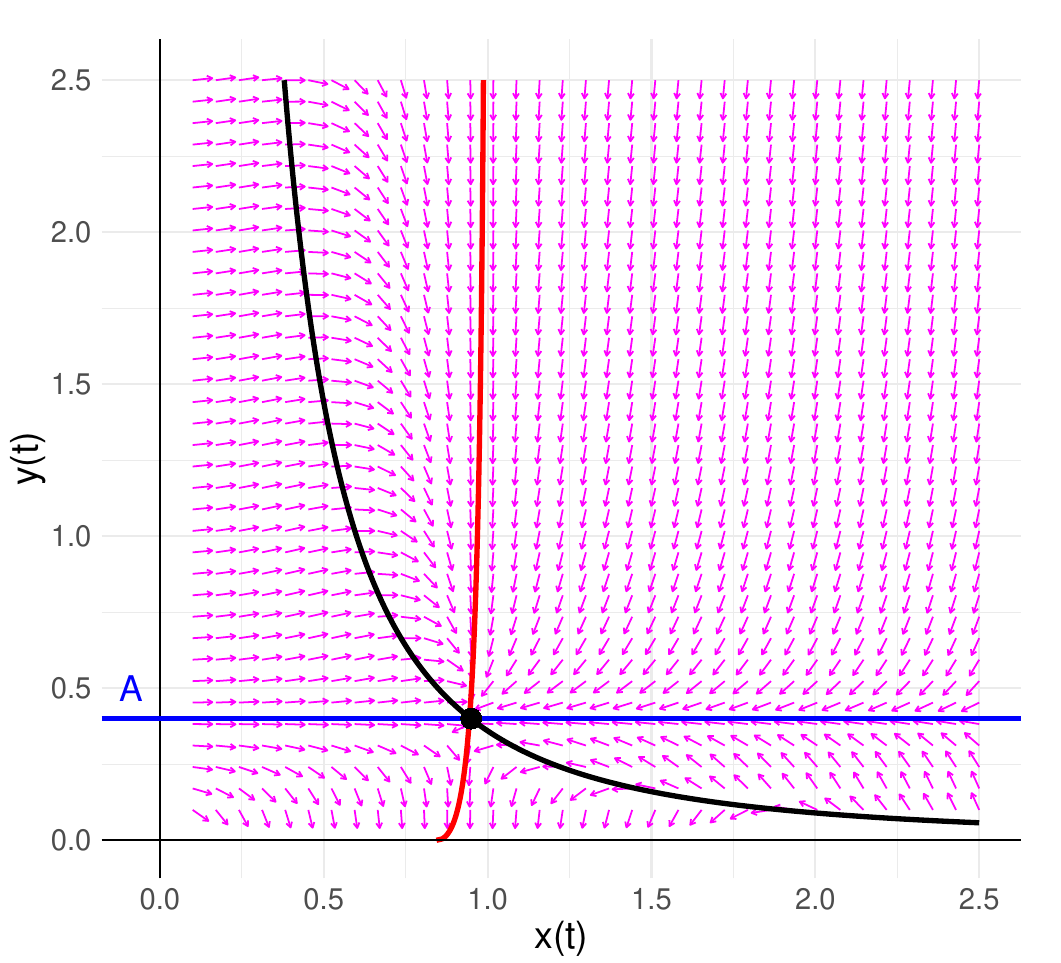}  \\
 (c) & (d) \\
\includegraphics[scale=0.45]{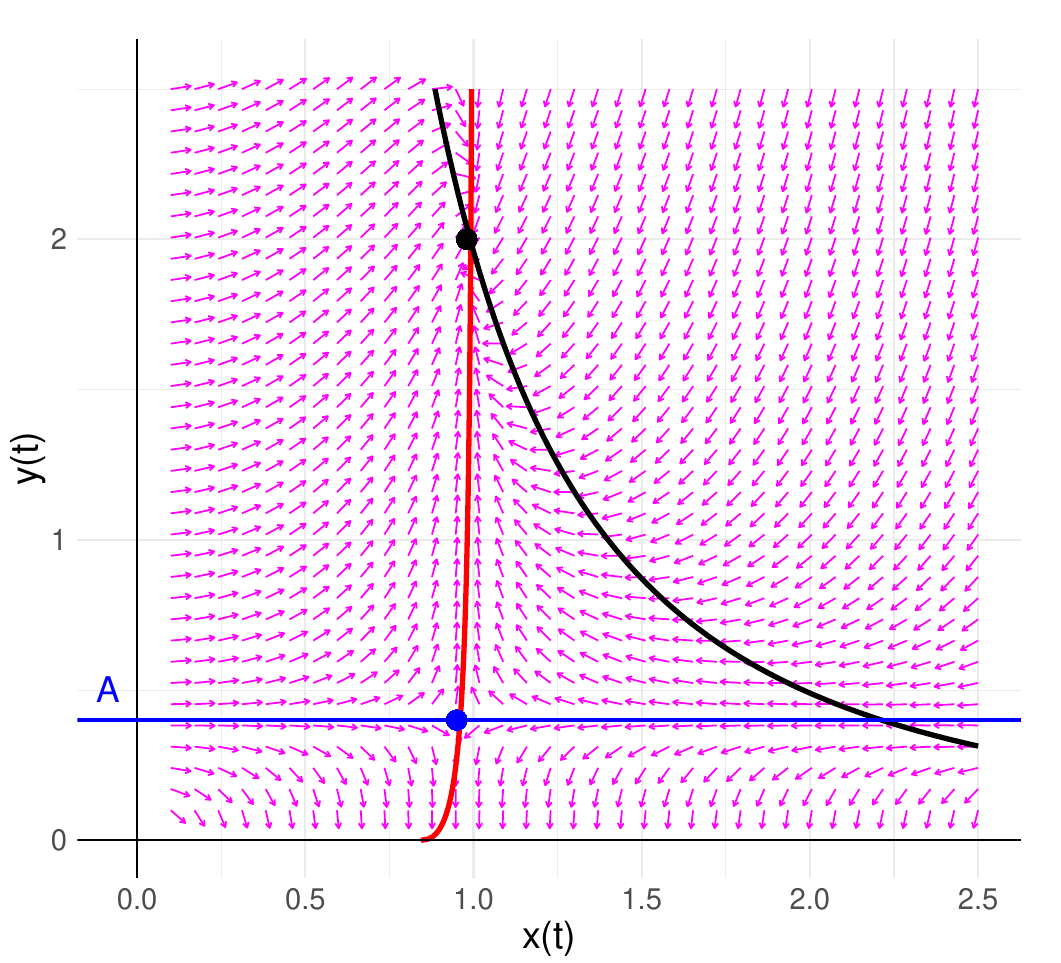} & \includegraphics[scale=0.45]{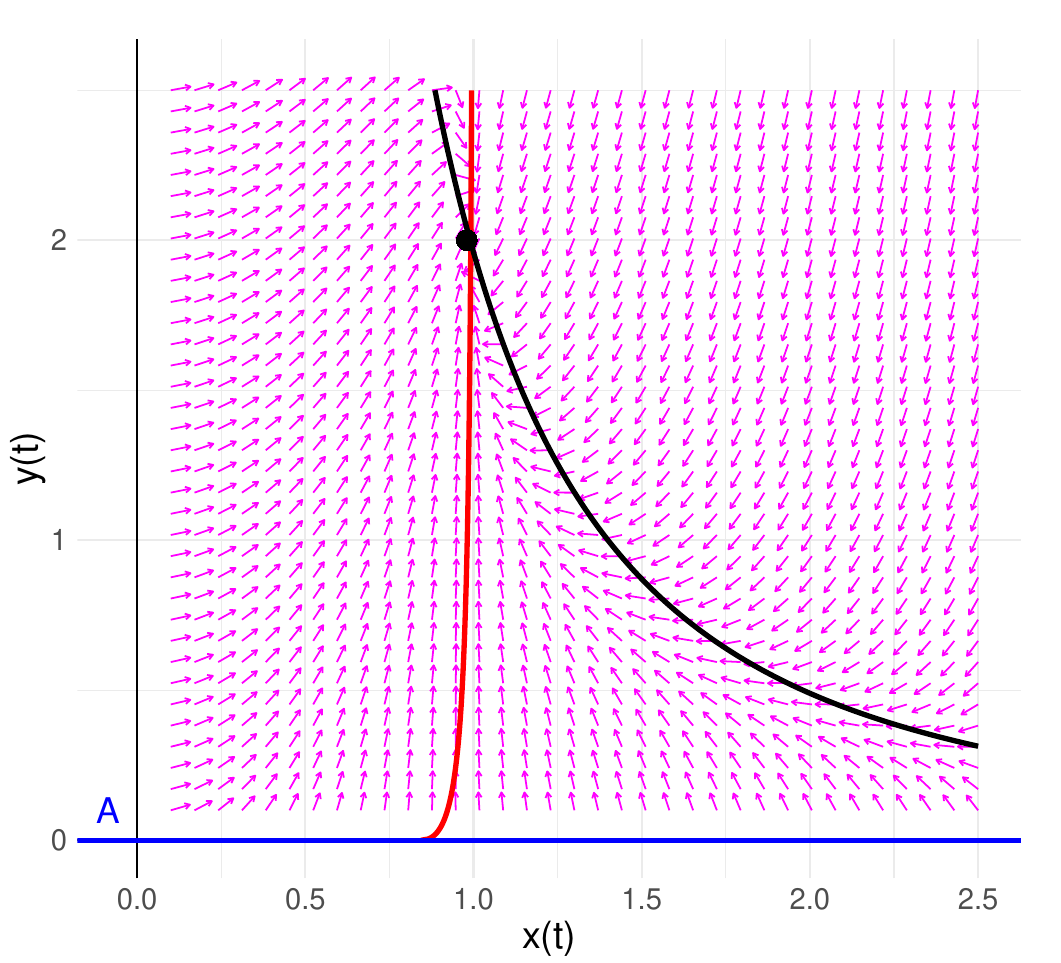} 
\end{tabular}
\end{minipage}} \caption{In Figure \ref{fig:fixedpoints} (a),  the interior  fixed point at $y=A$ is stable and the other is a saddle. We observe that a higher threshold $A$ increases the viability region of the system. In this case, not all weights below the Allee threshold lead to extension.  In figure (b),  both interior fixed points collide to create an unstable interior fixed point (a saddle). In Figure \ref{fig:fixedpoints} (c), the interior fixed point at $y=A$ is  a repeller  and the other is stable. Both Figures \ref{fig:fixedpoints} (b) and \ref{fig:fixedpoints} (c) are  illustration of the Allee effect where weights below the Allee threshold $y=A$ lead to extinction and those above converge the interior fixed point. We note also that  basin of attraction of the stable equilibrium is much larger in  Figure \ref{fig:fixedpoints} (c).  Figure \ref{fig:fixedpoints} (d) correspond to the Oja rule where $A=0$ and the interior fixed point in globally asymptotically stable.} \label{fig:fixedpoints}
 \end{figure}

\noindent The proof is provided  in Appendix \ref{App:A}.\\
\noindent   In the figures  (a), (b), and (c) below, we illustrate the results of the Theorem above.\\
   Figure \ref{fig:fixedpoints} (a) was obtained for $m = 0.01, u = 2.5, K = 0.4$ and $A = 1.7$.\\ 
     Figure \ref{fig:fixedpoints} (b) was obtained for $m = 2, u = 1.5, K = 0.4$ and $A = 0.4$.\\
 Figure \ref{fig:fixedpoints} (c) was obtained for $m = 2, u = 2, K = 0.7$ and $A = 0.4$.\\ 
  Figure \ref{fig:fixedpoints} (d) was obtained for $m = 2, u = 2, K = 0.7$ and $A = 0$.\\ 
We note that the blue and black curves represent respectively $y=A$ and $y=\left(\frac{uK}{x}\right)^2$,  pieces of the isocline $\ds \frac{dy}{dt}=0$, whereas the red curve represents the isocline $\ds \frac{dx}{dt}=0$ or $x=G(u\sqrt{y}+mx)$.

\subsubsection{Bifurcation analysis}
 Bifurcation analysis is important because, first, it explains how neurons and networks transition between different states; second, it provides a framework for understanding both normal and pathological brain dynamics; and third, it guides the design of interventions and technologies that modulate neural activity.
 The dynamics of the system depend on parameters \(A, K,u\),  and \(m\). Bifurcation analysis examines how the fixed points and their stability change as these parameters vary. For sake of self-containment, we start with a brief graphical review (see Figure \ref{fig:RevBifurcation} below) of the type of bifurcations under consideration in this paper. We used a generic parameter $\mu$ on the $x$-axis and the $y$-axis represent the fixed point of the system.
 \begin{figure}[htbp] 
    \centering
    \includegraphics[width=4.5in]{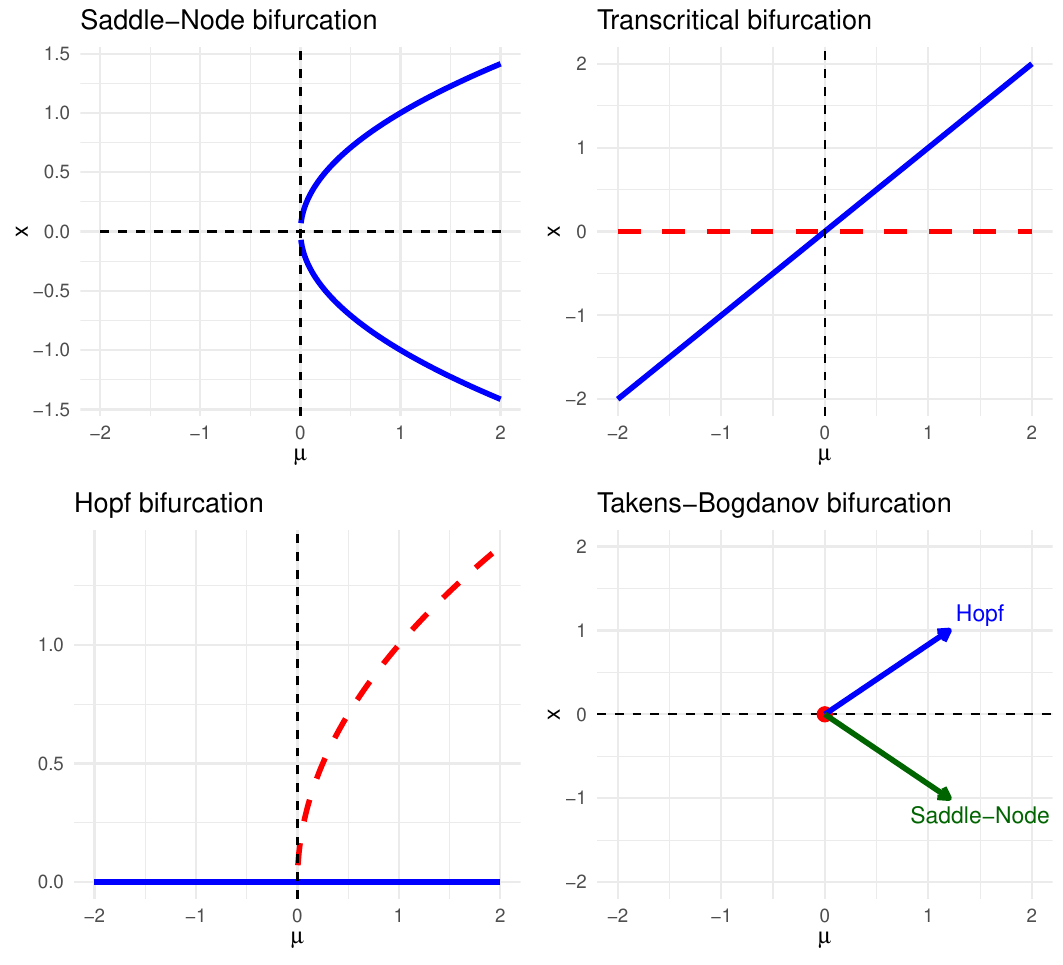} 
    \caption{For saddle node bifurcation, two fixed points (one stable, one unstable) collide when $\mu$ approaches 0. For transcritical bifurcation, two branches cross and exchange stability. For Hopf-bifurcation, a limit cycle appears as  $\mu$ crosses zero. Takens-Bogdanov bifurcation appears at  the intersection of saddle-node and Hopf bifurcations for $\mu=0$.}
    \label{fig:RevBifurcation} 
 \end{figure}
\noindent In the context of the Allee model, we have: 

 \noindent  {\bf Saddle-Node bifurcation}\\
 As \(u\) (pre-synaptic weight growth factor) decreases, the two fixed points --one stable, one saddle--can collide and annihilate each other, leading to extinction (see figure \ref{fig:fixedpoints} (b) above).
  
 \noindent {\bf Transcritical bifurcation} \\
 As \(A\) (weight threshold) decreases, a stable fixed point (viable weight) and an unstable fixed point (extinction state) may exchange stability (see Figure \ref{fig:fixedpoints} (a) and (c) above).

\noindent {\bf Hopf bifurcation}\\
 For certain combinations of \(u, m, K\), oscillations in \(x\) and \(y\) may arise, corresponding to periodic synaptic output-weight cycles. A Hopf bifurcation explains the emergence of rhythmic neural activity in brain regions such as the hippocampus and cerebral cortex. Oscillations in neural networks often arise from the interplay between excitatory and inhibitory activity. A Hopf bifurcation provides a framework for understanding how this balance influences system stability and rhythmic behavior. Mathematically,  a  Hopf Bifurcation occurs when a pair of conjugate eigenvalues $\lambda_{1,2}=\alpha \pm i\theta$ crosses the imaginary axis (that is, $\alpha=0$ and $\theta\neq 0$). In the Allee model case, this amounts to  $tr(x^*,y^*)\equiv trace(J)=0$ and $det(x^*,y^*)\equiv det(J)>0$, where $J$ is the Jacobian matrix  of system evaluated at the fixed point $(x^*, y^*)$. In this case, the eigenvalues are $\ds \lambda_{1,2}=\pm i\sqrt{-det(J)}$. We recall that the fixed points of the system are given as $(x^*=G(u\sqrt{y*}+mx^*), y^*=\left(\frac{uK}{x^*}\right)^2)$ and $(x_A^*=G(u\sqrt{A}+mx_A^*), y^*=A)$.  We state the following result  regarding Hopf bifurcation.
      
      \begin{thm}\label{thm2} Let $(x^*, y^*)$ be a fixed point of the  Allee model. 
      Put \[\lambda=1-\frac{A}{y^*}, \quad \beta=\frac{(uK)^2}{2m}\;.\]
    For $y^*=A$, the Allee model  does not possess a Hopf bifurcation at $(x^*=x_A, y^*=A)$.\\
    For  $y^*\neq A$,  the Allee model possesses a Hopf bifurcation at $(x^*=G(u\sqrt{y*}+mx^*), y^*=\left(\frac{uK}{x^*}\right)^2)$  if 
    \begin{itemize}
  \item  $\ds y^*<A$ and $\ds \tau_A<\xs<\sqrt{p_2}$\;, or 
   \item  $\ds y^*>A$ and $\ds \xs>\sqrt{p_2}\;,$
    \end{itemize}
   where \[ p_2=\frac{\beta}{2}+\sqrt{\frac{\beta^2}{4}+\frac{2\beta}{\lambda}}.\]
      \end{thm}
   \noindent   The proof can be found in Appendix \ref{App:B}.
      \begin{remark} This result shows in particular that when $A=0$ as in the Oja and Hebbian model, there is no Hopf bifurcation, thus showing that the Allee model has a richer dynamic than than the other two.
      \end{remark}
      \noindent In the  figures below, we used $u= 2.0, m=5, K= 2$.
\begin{figure}[H]
    \centering
     \centering  \begin{tabular}{cc}
   (a) & (b)\\
    \includegraphics[scale=.3]{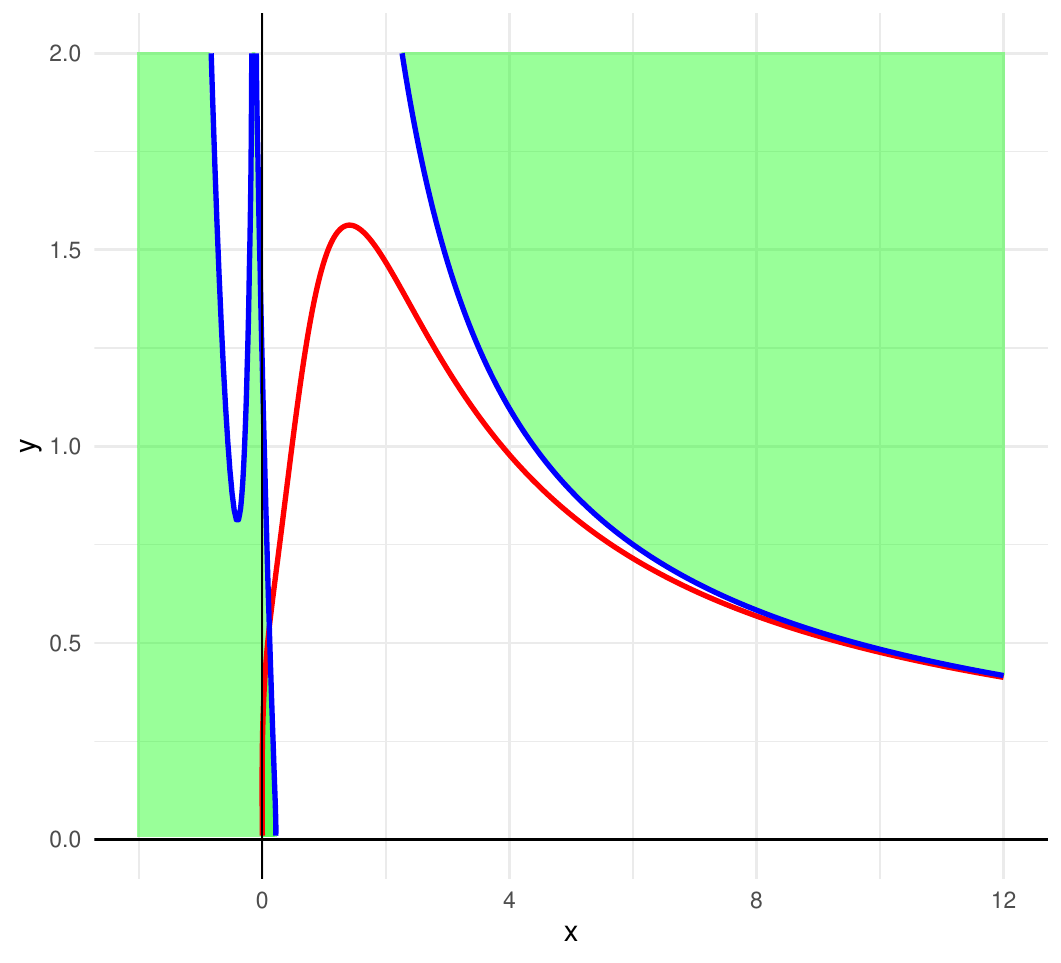} &   \includegraphics[scale=.3]{ 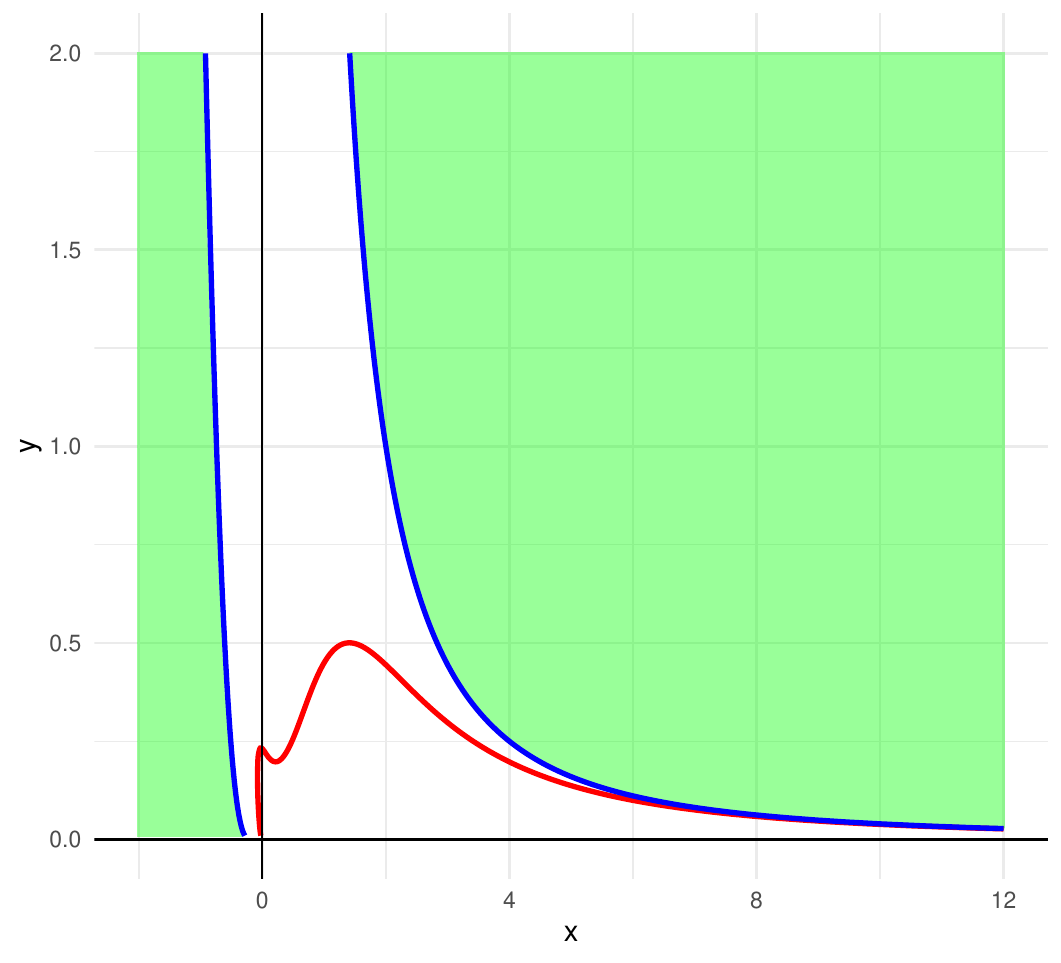}\\
       (c) & \\
        \includegraphics[scale=.3]{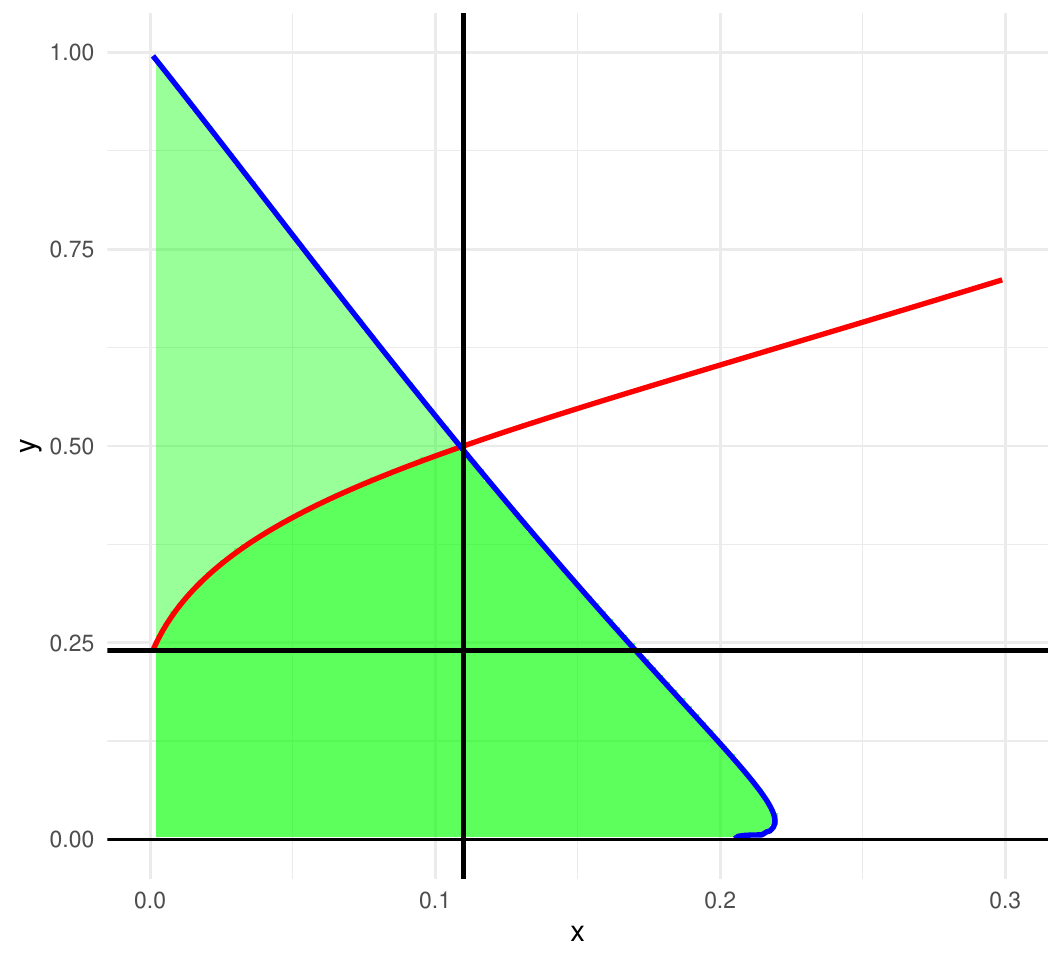}  & \\
      \end{tabular}
    \caption{The blue curve represents $det(x,y)=0$ whereas the red represents $tr(x,y)=0$ for $A=1$ in (a).  The green shaded region represents $det(x,y)>0$. We see that the part of the green region containing the red curve represents the Hopf bifurcation region for this model. In (b), we take $A=0$. we observe that there no part of the green region ever containing the red curve, a sign of the absence of a Hopf bifurcation. In (c), the estimated  region where $det(x,y)>0, tr(x,y)=0)$ bounded by the line $x\approx 0.11$ and $y\approx 0.23$.}
    \label{fig:TraceDeter}
 \end{figure}
 \noindent {\bf Takens-Bogdanov bifurcation} \\
 This type of bifurcation occurs when $tr(x,y)=det(x,y)=0$ and the Jacobian at that point is a nilpotent matrix.  In Figure \ref{fig:TraceDeter} (c) above,  we observe that at the intersection between the blue and red curves, we may have  a Takens-Bogdanov bifurcation. However, further analysis is needed to prove the nilpotence of the Jacobian.
 \subsubsection{Memory Storage }\label{MemoryStorage} 
 Memory storage refers to the encoding, retention, and retrieval of information, involving both structural and functional changes from synapses to brain circuits. Key types of memory include short-term, long-term, working, episodic, and procedural memory. These processes rely on mechanisms such as synaptic plasticity, Hebbian learning, and systems-level consolidation. Critical brain regions include the hippocampus (for declarative memory), amygdala (emotional memory), prefrontal cortex (working memory), cerebellum and basal ganglia (procedural memory), and neocortex (long-term storage). Understanding these systems is crucial, as disorders like amnesia, Alzheimer’s, PTSD, and learning disabilities directly impact memory function. For the Allee model, key features relevant to memory storage include:
 
\noindent {\bf Fixed points} \\
The system's fixed points represent stable states that can ``store" information about initial conditions.  The presence of multiple stable equilibria means the system retains memory of its initial conditions. Whether the system evolves to extinction 
$(x=0)$ or to a positive equilibrium depends on where it starts. This property enhances long-term memory, as the system ``remembers'' the basin of attraction from which it originated
\noindent {\bf Attractors} \\
Attractors often represent stable neural states, such as specific patterns of activity in neural populations, which can correspond to cognitive or behavioral states, memories, or motor commands. 
Stable nodes or limit cycles act as memory by determining long-term system behavior. Therefore, small perturbations may be insufficient to move the system out of an attractor basin (see Figures \ref{fig:fixedpoints} (a), (c),  (d)), that is, memory is resistant to noise.
\noindent {\bf Time delays} \\
The sigmoid function $G$ introduces nonlinearity that can emulate memory-like behavior due to threshold effects.  Let us explain mathematically a mechanism that mimics synaptic potentiation, that is, weak activations are discarded and only meaningful patterns are remembered. The sigmoid $G(z)$ has asymptotes at  0 and 1, with a soft threshold around $z=0$. Now consider  the equation  $\frac{d x}{dt} \vspace{0.25cm}=-x+G(u\sqrt{y}+mx)$. 
\begin{itemize}
    \item[(a)] Forgetting: 
    If $z=mx + u\sqrt{y}<0$, then $G(z)\approx 0$, therefore $\frac{d x}{dt} \approx -x\implies x(t)\approx x(0)e^{-t}$.
This means that small perturbations in $x$ or $y$ that do not push the input $mx + u\sqrt{y}$ above the sigmoid's threshold decay quickly. 
    The system ``forgets'' these disturbances because the nonlinear response is near zero and does not sustain these perturbations.

    \item[(b)] Retention: If $z=mx + u\sqrt{y}\gg0$, then $G(z)\approx 1$,  therefore $\frac{d x}{dt} \approx -x+1 \implies x(t)\approx 1-(x(0)-1)e^{-t}$. This means that perturbations that push $mx + u\sqrt{y}$ above the threshold cause $G(z)$ to  increase rapidly  toward 1. In this regime, the system exhibits persistence. The nonlinearity in $G(z)$ allows these larger perturbations to be ``remembered'' as they are amplified by the feedback dynamics in the system.
\end{itemize}
While above we focused on the output neuron $v$, we can also mimic  synaptic potentiation by regulating weights through the Allee threshold. Neurons with synaptic weights below the Allee threshold are discarded, whereas those above the threshold are preserved or retained. 
\subsubsection{Sensitivity to parameters}
\begin{figure}[H]
\centering\begin{tabular}{c}
\includegraphics[scale=0.5]{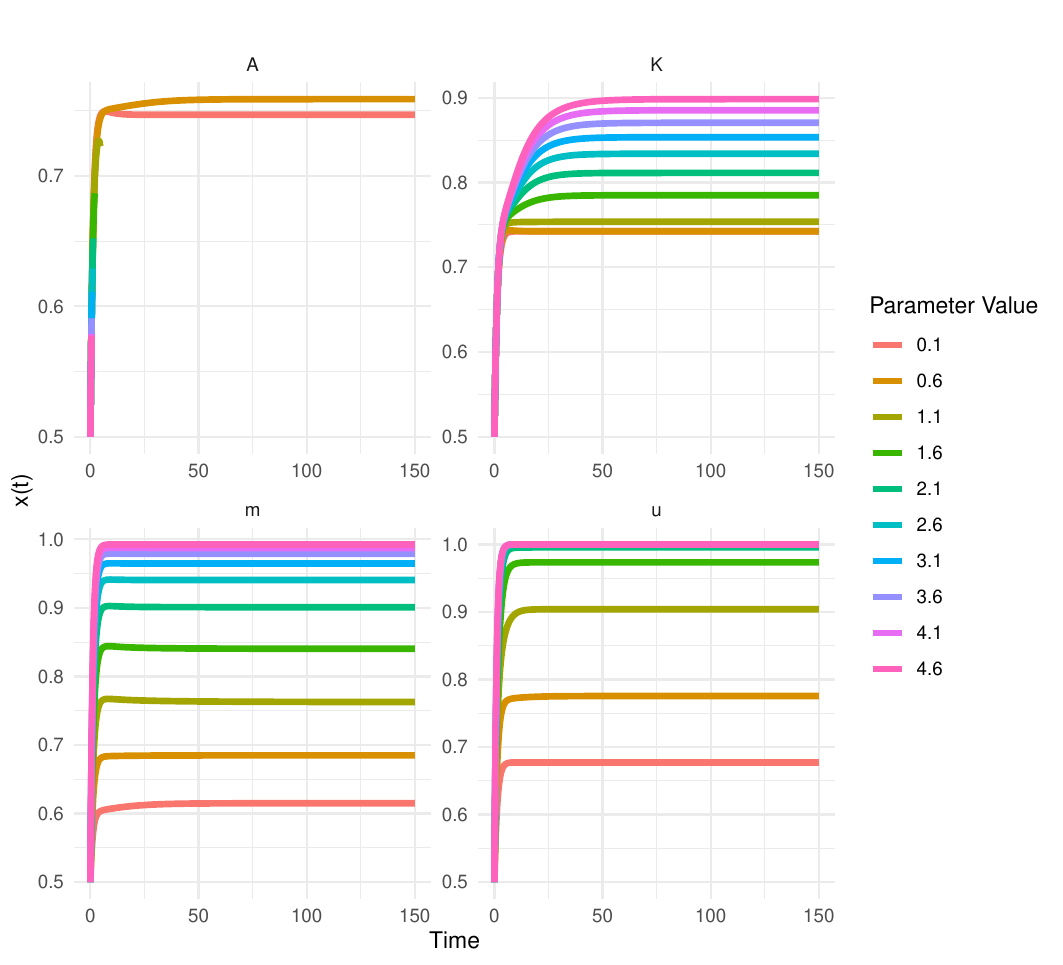}\\
\includegraphics[scale=0.5]{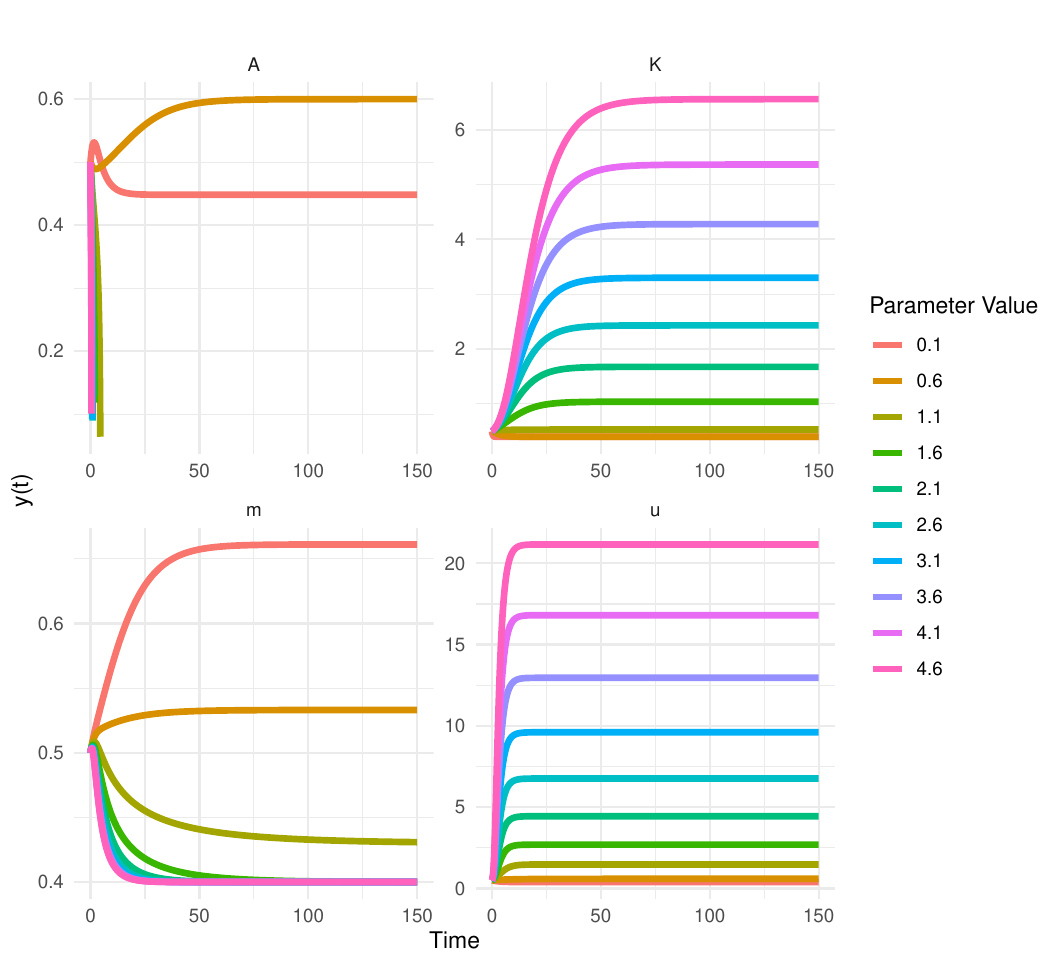}\\
\end{tabular}
\caption{In the first plot above, we  represent the sensitivity of $x(t)$ to parameters $A,m$ and $u$ and that of $y(t)$ in the second plot below. Higher $A$ increases the threshold for $y$, reducing the likelihood of feedback effects. Low $A$ may destabilize the system by allowing strong feedback even for small perturbations.  Higher $K$ induces higher growth in weights, which may lead to instability in the system.  Higher $m$ increases the influence of $x$ on $G$, leading to stronger memory effects and delayed decay. Higher $u$ strengthens the influence of $y$ on $G$, increasing the system's responsiveness to $y$. Small $u$ values may lead to weaker coupling between $x$ and $y$. }
\label{fig:SensitivityAnalysis}
\end{figure}

Sensitivity to model parameters like $A, K, m$, and $u$ must be analyzed to ensure robustness across a range of biologically plausible scenarios.
Since the roles of $A$ and $K$ have been discussed before, we focus on the others:\\
\noindent  (a) Parameter $m$:
\begin{itemize}
\item $m$ controls the contribution of $x$ to the argument of the sigmoid function.
\item A higher value of  $m$ can lower the activation threshold for $G$, promoting faster transitions and enhancing a more robust memory storage, as previously discussed.
\end{itemize}
\noindent  (b) Parameter $u$ 
\begin{itemize} 
\item $u$ modulates the growth term and the interaction between $x$ and $y$.
\item A higher $u$ can strengthen the positive feedback loop, reinforcing stable equilibria and improving memory retention.
\end{itemize}
An extended simulation (Figure  \ref{fig:SensitivityAnalysis})  illustrates model performance across a broader range of parameter values
 including stress tests under variable noise and initial states. The results show that the model retains qualitative robustness across moderate fluctuations in $A, m$, and $u$. When $A$  is varied, we fix $m = 1, u = 0.5, K = 1$.  When $K$  is varied, we fix $m = 1, u = 0.5, A = 0.4$. When $m$ is varied, we fix $ u = 0.5, A = 0.4, K = 1$. When $u$ is varied, we fix $m = 1, A = 0.4, K = 1$. Each parameter is varied for ten equidistant  values  from 0.1 to 4.6. The results show that the model retains qualitative robustness across moderate fluctuations in $A,K m$, and $u$.
\subsubsection{Pattern overlap}
Pattern overlap is a key metric in evaluating the effectiveness of neuroscience learning rules, as it quantifies how accurately a neural system can retrieve stored information from partial or noisy cues. High overlap indicates successful pattern completion and reflects the robustness of associative learning mechanisms. It also serves as a functional measure of memory retrieval fidelity, helping to compare the performance of different synaptic plasticity models, such as Hebbian, STDP, or Allee-based rules. By capturing how well a system converges to learned attractors under real-world variability, pattern overlap provides critical insights into the reliability and efficiency of biological and artificial memory systems.
To evaluate pattern retrieval in the Allee model, we assess whether the system's dynamics enable recovery of initial states or behavioral patterns via its attractors, fixed points, and stability characteristics. If a target pattern corresponds to a specific fixed point $(x^*,y^*)$,  the overlap at time $t$   is defined as
\[\text{Overlap}(t)=1-\frac{\sqrt{(x(t)-x^*)^2+(y(t)-y^*)^2}}{\sqrt{x^{*2}+y^{*2}}}\;.\]
This measure ranges from $0$ to $1$, where $1$ indicates perfect retrieval of the fixed point and $0$ signifies no correlation with it.  In Figure \ref{fig:Overlap} below, we illustrate the pattern overlap capability of the Allee model  for   seven different  initial conditions: $(x_0,y_0)\in \set{(0.1, 0.2), (0.3, 0.5), (0.6, 0.8), (0.9, 1.2), (1.5, 1.8),  (0.1,4),(2,0.1)}$ and  for $u=1, m=0.5, \alpha=1, K= 2$ and  $A= 0.4$. We considered time steps $t=0,1, \cdots, 20$. The results show that trajectories with initial weights below the Allee threshold exhibit zero overlap, which corresponds to extinction and thus no possibility of memory retrieval. Other trajectories with weight above the Allee threshold have  overlap equal to 1. This corresponds to convergence to the stable equilibrium and demonstrating that the system can retrieve stored patterns over time. In conclusion, these results confirm that the Allee model accommodates both memory retention and forgetting. The Allee effect from an ecological perspective corresponds to a situation where decay towards zero is irreversible. In memory settings, this  can be thought of as a mechanism of irretrievable memory. 
\begin{figure}[H]
\centering \includegraphics[scale=0.48]{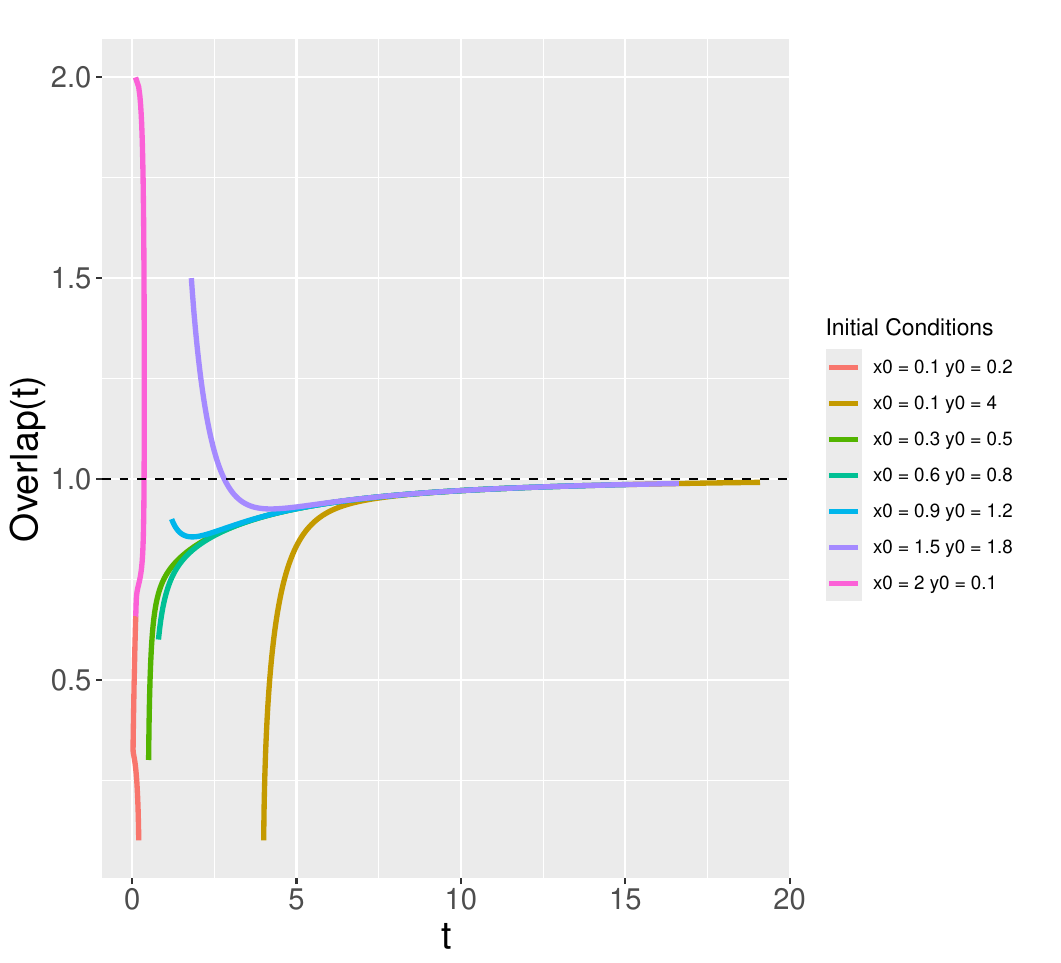}
\caption{Overlap pattern retrieval as a function of $t$ in the Allee Model. There are seven trajectories represented by their initial starting points $(x_0,y_0)$, for model parameters $u=1.0, m=0.5,\alpha=1, K= 2.0$ and $A= 0.4$. Two starting points (magenta and brown) with low values of $y$ (0.1 and 0.2) have zero  trajectory overlap over time, whereas the rest  have an  overlap of 1  over time.  Low values of $y$, especially below a low Allee threshold $A$ correspond dynamically to extinction, thus no convergence to a non-trivial fixed point. This is similar to what we observed  in the  Figures \ref{fig:fixedpoints} (b) and  \ref{fig:fixedpoints} (c) above.  In particular, it highlights  the influence of $y=w^2$, the length of weight  in the model. }
\label{fig:Overlap}
\end{figure}
\subsection{Multiple  layers and multiple post-synaptic neurons model} \label{sec:4}
The analysis of the single-layer neuron model highlighted the mathematical challenges encountered in analyzing model \eqref{eqn:Alleerule1}. Recall that in \cite{Kwessi2022}, the version of the model with a linear gain function was shown to regulate unbounded growth, preserves synaptic  normalization, avoids blow-up  at lower initial values, and  induces competition between weights. In the following sections, we investigate whether the inclusion of multiple layers affects the model's  robustness to noise.
\subsubsection{Associative memory and noise robustness}
Associative memory refers to the brain’s ability to store and retrieve information based on learned associations. It allows a full pattern to be retrieved from partial or noisy input. There are two main types: auto-associative memory, which retrieves a complete pattern from a partial cue, as in Hopfield network (\cite{Hopfield1984}), and hetero-associative memory, which links different types of patterns, such as associating a name with a face. To define   the associative memory of a network related to the Allee model above, we consider a network consisting of $L$ layers, $N_u$ pre-synaptic neurons, $N_v$ post-synaptic neurons,  synaptic weight block matrix ${\bf W}=(W_{ij}^{k\ell})$, initially set to zero, and where $W_{ij}^{k\ell}$ represents the strength of the connection between neuron $i$ (pre-synaptic) on layer $k$ and neuron $j$ (post-synaptic) on layer $\ell$. 
%The Allee learning rule extends the standard Hebbian rule by introducing nonlinear interactions and normalization terms that constrain synaptic weight growth.
 Therefore, for each pattern $({\bf u, v})$, the update rule is 
\[\Delta  W_{ij}^{k\ell}= v_j^{\ell}( u_i^{k}-K^{-1} W_{ij}^{k\ell} v_j^{\ell})\left(1-A /({\bf W}^T{\bf W})_{ij}^{k\ell}\right)\;,\]
where $A=0, K=\infty$ for the Hebbian rule, $A=0, K>0$ for the Oja  rule, and $A> 0, K>0$ for the Allee rule.  
This rule introduces a multiplicative normalization factor that depends on the squared weight norm.
We note that element-wise,  $\ds {\bf W}^T{\bf W}$ yields a matrix whose entries are given by
\[\sum_{m=1}^{N_v}W_{jm}^{k\ell}W_{im}^{k\ell},\quad \textrm{$1\leq i,j\leq N_u$ and $1\leq k,\ell\leq L$}\;.\]
For the Spike-Timing Dependent Potentiation (STDP) rule,  we  will consider five different formulations for $\Delta  W_{ij}^{k\ell}=F[(\Delta)_{ij}^{k \ell}]$ given in Table \ref{tab:stdp_rules} below.
\begin{table}[H]
\centering
\begin{tabular}{|l|l|}
\hline
\textbf{STDP Type} & \textbf{Weight Update Rule} $\boldsymbol{\Delta W }$ \\
\hline
\textbf{Pair-based} & 
$\begin{cases}
B_+ e^{-\Delta t / \tau_+}, & \Delta t > 0  \vspace{0.25cm}\\
-B_- e^{\Delta t / \tau_-}, & \Delta t < 0
\end{cases}$ \\
\hline
\textbf{Weight-dependent} & 
$\begin{cases}
B_+ (1 - w) e^{-\Delta t / \tau_+}, & \Delta t > 0  \vspace{0.25cm}\\
-B_- w e^{\Delta t / \tau_-}, & \Delta t < 0
\end{cases}$ \\
\hline
\textbf{Additive/Multiplicative} & 
$\begin{cases}
B_+ (1 - w) e^{-\Delta t / \tau_+}, & \Delta t > 0  \vspace{0.25cm}\\
-B_- w e^{\Delta t / \tau_-}, & \Delta t < 0
\end{cases}$ \\
\hline
\textbf{Power-law} & 
$\begin{cases}
B_+ (1 - w)^\gamma e^{-\Delta t / \tau_+}, & \Delta t > 0 \vspace{0.25cm}\\
-B_- w^\gamma e^{\Delta t / \tau_-}, & \Delta t < 0
\end{cases}$ \\
\hline
\textbf{Continuous-time (differential)} & 
$B \cdot \frac{\Delta t}{\tau_+^2} e^{-|\Delta t| / \tau_+}$ \\
\hline
\end{tabular}
\caption{Formulations of five different STDP learning rules.}
\label{tab:stdp_rules}
\end{table}
\noindent Here,  $B_{-1}>0$ and $B_+>0$ represent respectively pre-and post synaptic  potentiation amplitudes, and  $(\Delta t)_{ij}^{kl}=t_j^{\ell}-t_i^{k}$ represents the time difference between post-and pre-synaptic spikes on layers $\ell$ and $k$.  The parameters $\tau_-$  and $\tau_+$ are the decay time constants governing how quickly the weights diminish over time and the update rule reflects memory accumulation at discrete intervals $(\Delta t)_{ij}^{kl}$. $\gamma$ is a positive constant in $(0,1)$ and $B$ is a positive constant representing the potentiation amplitude in the continuous case.
A stored pair $({\bf u,v})$ is said to be retrieved if,  starting from a noisy version ${\bf u}^{\textrm{noisy}}$, the dynamics drive the system toward a stable fixed point or orbit close to ${\bf v}$.
Pattern retrieval is evaluated by introducing controlled noise to stored patterns, defined as
\begin{equation}\label{eqn:noise}
{\bf u}^{\text{noisy}}=\begin{cases}
-{\bf u_a}, & \text{if $a$ belongs to the noisy indices}\\
{\bf u_a}, & \text{otherwise}
\end{cases}\;.
\end{equation}
Now we suppose that we have $P$ patterns $({\bf u}_{\mu}, {\bf v}_{\mu})$ for $\mu=1, 2, \cdots, P$.  We then apply iterative updates to recover the original pattern as:

\noindent (1) Weight update:
\[
 W_{ij}^{k\ell}]_{\text{new}}\leftarrow  W_{ij}^{k\ell}]_{\text{old}}+\eta  \Delta  W_{ij}^{k\ell}\;.
\]
\noindent (2) State vector update: the retrieval process iteratively updates the state vector ${\bf s}$ based on the current weights:
\begin{eqnarray*}
{\bf {\bf v}^{(t)}}&=&\text{sign}({\bf u^{(t)}}{\bf W}), \quad {\bf u^{(0)}}={\bf u}^{\text{noisy}}\;.\\
\end{eqnarray*}
where ${\bf u^{(t)}}\in \set{+1,-1}^{N_u}$ and $\text{sign}(x)=+1$ if $x>0$ and $\text{sign}(x)=-1$ otherwise. The iteration proceeds until convergence to  a fixed point, that is, ${\bf v^{(t+1)}}={\bf v^{(t)}}$. The process makes the original pattern an attractor. 
\noindent Retrieval accuracy is calculated as the proportion of correctly recovered output neurons, and is computed as\[ \text{Accuracy}=\frac{1}{LN_v}\sum_{i=1}^{N_v} {\bf1}\left({\bf v}_i^{\text{retrieved}}={\bf v}_i^{\text{original}}\right)\;,\]
where ${\bf 1}(x)$ is the indicator function.
\noindent The  noise level is given as  \[\sigma=\frac{1}{LN_v}\sum_{i=1}^{N_v} {\bf 1}\left({\bf v}_i=-{\bf v}_i\right)\;.\]  
This represents the fraction of neurons flipped out of a total of $LN_v$ (see equation \eqref{eqn:noise} above).
 In Figure \ref{fig:patterns} below, we first illustrate the retrieval accuracy of the Allee model, showing a sample  input, noisy, and retrieved pattern. The parameters  used are $\sigma=0.3,\eta=0.01, LN_v=250, P=150, A=2$, and $K=1$. A visual inspection shows that  the majority of the pattern  is retrieved. The mean accuracy percentage across all patterns is around 54\%. 
 \begin{figure}[H]
\centering\includegraphics[scale=0.7]{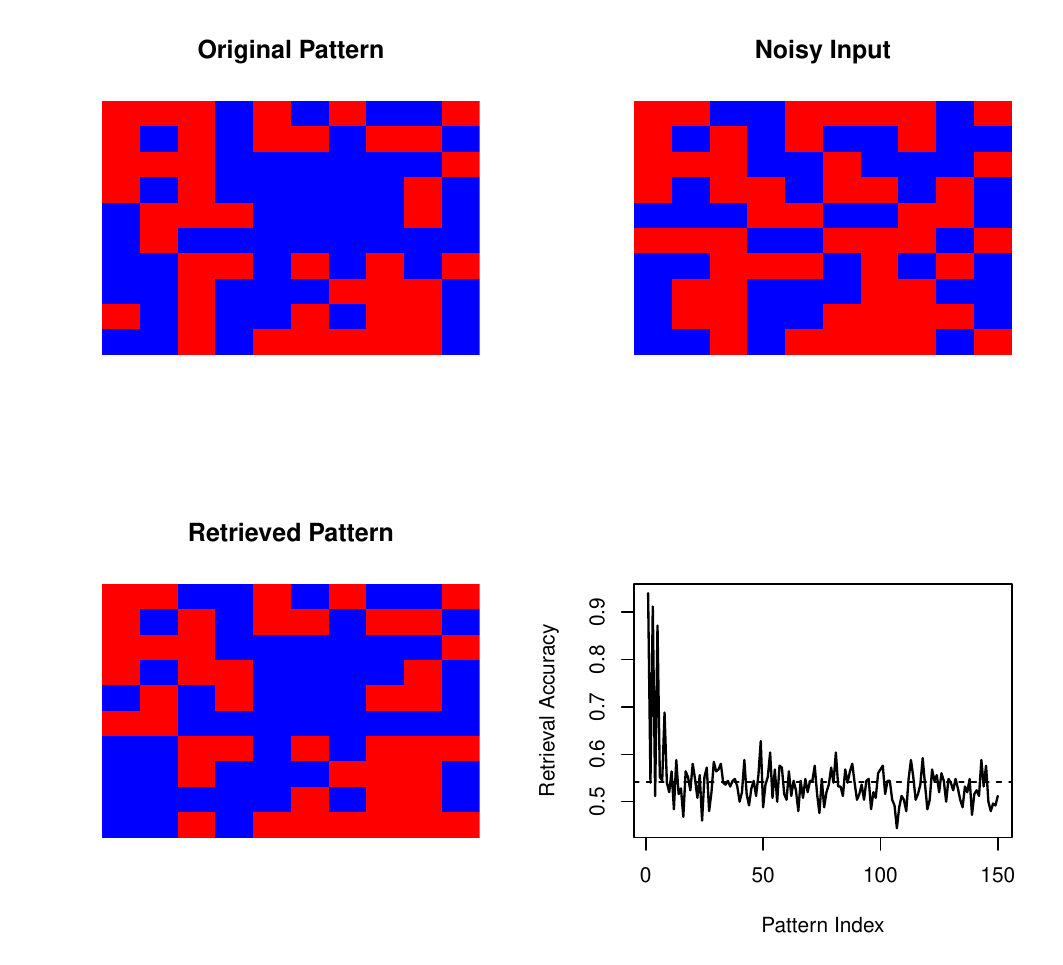}
\caption{The figure represents a comparison between original and retrieved pattern using an Allee update rule given above. Visual inspection shows relative similarity between the original and retrieved patterns. This demonstrates that the Allee model performs acceptably  well in pattern retrieval tasks. In the bottom right panel, the dashed line represents the mean accuracy ($\approx 54\%$) across all patterns.}
\label{fig:patterns}
\end{figure}
%\subsection{ and network size}
In Figure  \ref{fig:NoiseRobustness_Comparison} below, we compare the mean accuracy across all patterns   as a function of noise level for    the Allee, Hebbian, Oja, and the five STDP-types rules introduced above. The parameters used are:   $N_u=N_v=25, L=5, P=10, A=1, K=5$, and and $\eta=0.01$. For the STDP models, we choose  $B=B_+=0.01,  B_-=0.012, \tau_+=\tau_-=20, \gamma=0.7$ and   $\Delta t = 0.1$. 
\begin{figure}[H]
\centering \begin{tabular}{c}
 \includegraphics[width=5in]{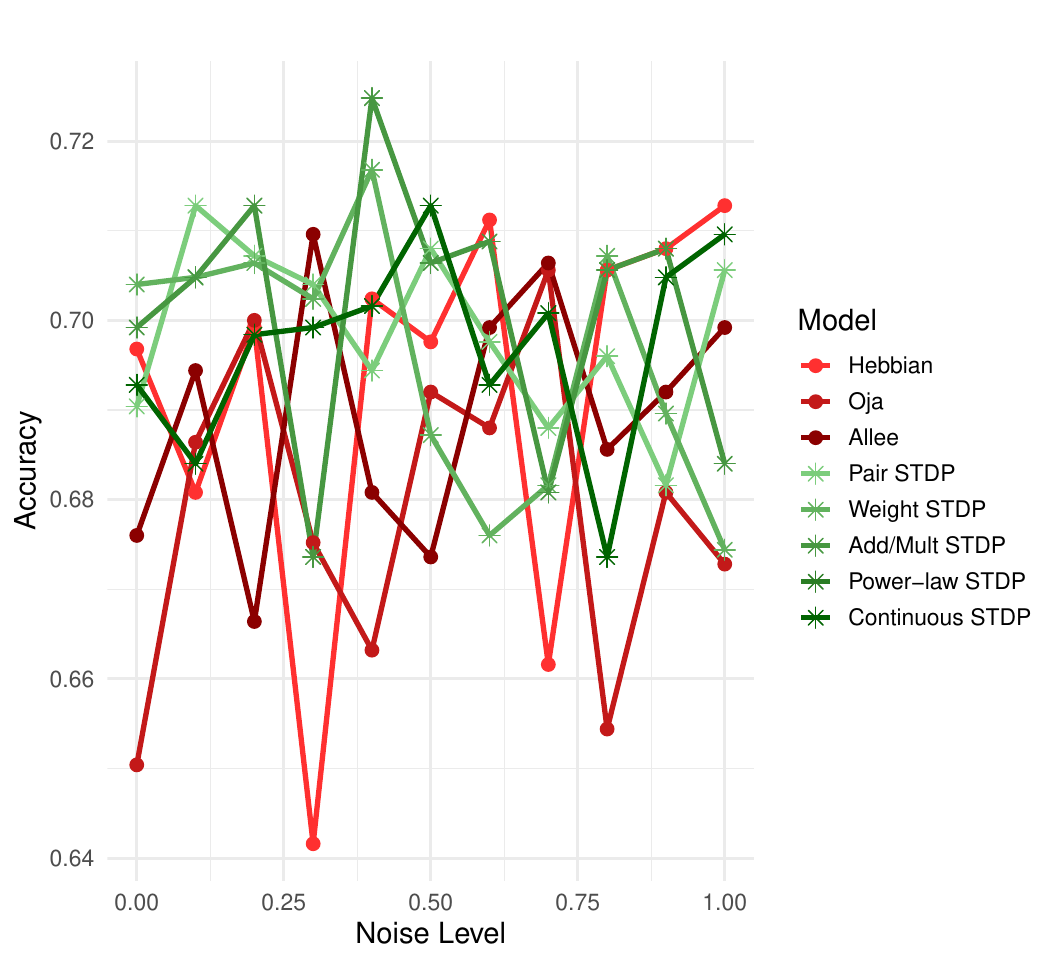}\\
\end{tabular}
\caption{ In this  figure, we compare the retrieval accuracy of  Hebbian (light red), Oja (moderate red),  Allee (dark red) and five STDP model variants (shades of green). STDP rules overall have  better retrieval accuracies which  on average  are at  around 70\%.  }
\label{fig:NoiseRobustness_Comparison}
\end{figure}
From Figure \ref{fig:NoiseRobustness_Comparison}  above, we can observe that  the Hebbian, Oja, and Allee models have a somewhat similar robustness to noise, with the Allee model occasionally being superior to the others. However its  accuracy is less than  that of STDP models in general. We note that  the simulations that led to  the figures above use a certain degree of randomness, therefore, repeating them will not yield the same curves. However, as aforementioned, the Allee model  also possesses other advantages like synaptic normalization and competition between neurons. A comparative analysis of these models is given in Table  \ref{tab:comparison} below:
\begin{table}[H]
\centering
   \resizebox{1\textwidth}{!}{\begin{minipage}{1.3\textwidth}
   \centering\begin{tabular}{|@{}l|l|l|l|l|@{}}
\toprule
\textbf{Feature}                & \textbf{Hebbian} & \textbf{Oja} & \textbf{STPD} & \textbf{Allee} \\
\midrule
Weight Regulation               & No regulation         & Normalization        & Spike timing dependent       & Nonlinear suppression       \\
Temporal Dynamics               & None                  & None                 & Captures spike interactions  & None                       \\
Noise Robustness                & Moderate                   & Moderate             & High                        & High                       \\
Biological Plausibility         & High                  & Moderate             & Very High                   & Moderate                   \\
Parameter Sensitivity           & Low                   & Moderate             & High                        & High                       \\
Stability                       & Poor                  & High                 & High                        & Moderate (tunable)         \\
Complexity                      & Low                   & Moderate             & High                        & High                       \\
Memory Capacity                 & Low                   & Moderate             & High                        & High                       \\
Realistic Synapse Behavior      & Limited               & Limited              & High                        & Critical threshold behavior \\
\bottomrule
\end{tabular}
\end{minipage}}
\caption{Comparison of plasticity models based on selected features.}
\label{tab:comparison}
\end{table}
\subsubsection{Incorporating time: temporal dynamics}\label{TemporalDynamics}
The Allee, Hebbian, and Oja models lack temporal dynamics--time-dependent changes in neural activity, connectivity, and behavior that  underlie how the brain processes information, encodes memories, controls movements, and adapts to new stimuli or environments. This may  account for their reduced resilience to noise compared to the STDP models.  Temporal dynamics can be  considered through two main mechanisms:  oscillatory dynamics and eligibility traces that may work simultaneously   to enable precise temporal processing. Oscillatory dynamics enable the brain to process, encode, and integrate time-dependent information  across various scales  while eligibility traces preserve a decaying memory of prior activity, enabling time-dependent associations. Studies have shown the importance of temporal dynamics through eligibility of traces: for instance   in \cite{Shindou2019}, it was   demonstrated that dopamine-dependent synaptic changes with delays up to seconds after spike event, and in \cite{SchultzDayanMontague1997},  the authors  showed that dopaminergic neurons signal reward prediction errors--a key factor in reinforcement learning. We incorporate   temporal dynamics in the Allee model as follow:
\begin{equation}\label{eqn:modifiedAlleeModel2}
\begin{cases}
\ds \tau_{_{\bf v}} \frac{d \bf v}{dt} \vspace{0.25cm}&=-{\bf v}+G({\bf W, M, u,v})+\kappa E_{\bf v}(t)\\
\ds \tau_{_{\bf W}} \frac{d {\bf W}}{dt} &={\bf v}^T\left({\bf u}-K^{-1}{\bf  W}{\bf v}\right)\left({\bf 1}-A ({\bf W}^T{\bf W})^{-1}\right)+\lambda E_{\bf W}(t)
\end{cases}\;,
\end{equation}
where  $E_{\bf v}(t) =e^{-\Delta t / \tau_1},  E_{\bf W}(t) =   e^{-\Delta t / \tau_2}$ are eligibility of trace,  $\tau_1,\tau_2>0$ are some decay constants, and $\Delta t=t-t_{\text{pre}}$ is the time difference since pre-synaptic spiking. Here $\kappa$  (post-synaptic potentiation amplitude) controls the influence of the eligibility trace $E_{\bf v}$ which retains a memory of recent ${\bf v}(t)$-activity and decays exponentially over time, while $\lambda$ (weights potentiation amplitude) scales the effect of the eligibility trace $E_ {\bf W}$ which retains a memory of recent ${\bf W}(t)$-activity. The proposed model introduces greater dynamic complexity, but, if computational challenges are addressed, the model greatly improves associative memory and pattern retrieval. It also retains critical properties of nonlinear feedback and robustness. The presence of the Allee threshold makes it suitable for systems exhibiting bi-stability or multi-stability, mimicking biological networks with discrete activation states.  The weight update is given as 
\[
 W_{ij}^{k\ell}]_{\text{new}}\leftarrow W_{ij}^{k\ell}]_{\text{old}}+\Delta W_{ij}^{k\ell}\;,\]
 where 
 \[
 \Delta W_{ij}^{k\ell}=\begin{cases}
v_j^{\ell}( u_i^{k}-K^{-1}  W_{ij}^{k\ell} v_j^{\ell})\left(1-A /({\bf W}^T{\bf W})_{ij}^{k\ell}\right)+ \kappa e^{-(\Delta t)_{ij}^{kl}/\tau_1} & \textrm{if $(\Delta t)_{ij}^{kl}>0$ }\\
 v_j^{\ell}( u_i^{k}-K^{-1}  W_{ij}^{k\ell} v_j^{\ell})\left(1-A /({\bf W}^T{\bf W})_{ij}^{k\ell}\right)+\lambda e^{-(\Delta t)_{ij}^{kl}/\tau_2} & \textrm{if $(\Delta t)_{ij}^{kl}<0$ }
 \end{cases}\;.
\]
Here,  $(\Delta t)_{ij}^{kl}=t_j^{\ell}-t_i^{k}$ represents the time difference between post-and pre-synaptic spikes on layers $\ell$ and $k$. 

\begin{figure}[H] 
   \centering
   \begin{tabular}{c}
   \includegraphics[scale=0.7]{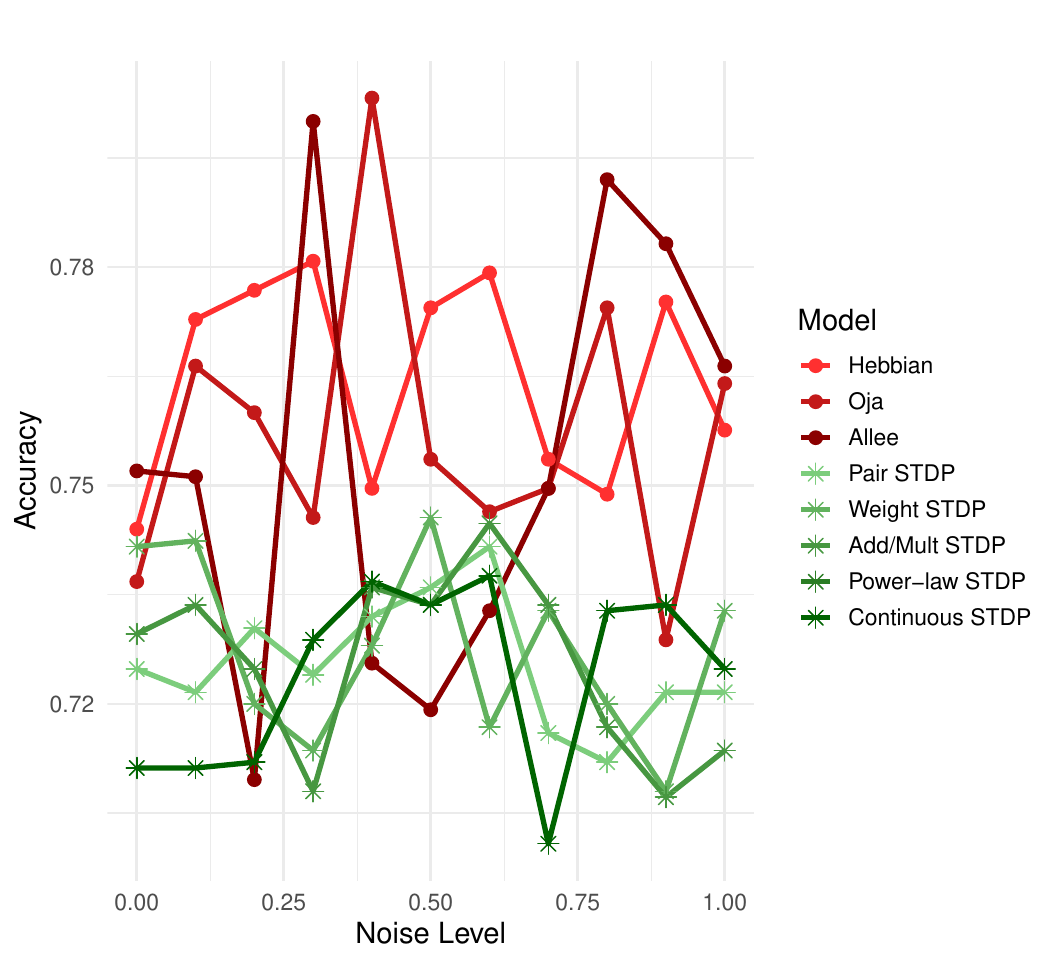} 
   \end{tabular}
   \caption{This plot shows the retrieval accuracy of  Hebbian (light red), Oja (moderate red),  Allee (dark red) with temporal dynamics and five STDP model variants (green). Clearly the accuracy  the first three model is improved by an addition of temporal dynamics.}
   \label{fig:RetrievalAccu_Noise_Temp_Dyn}
\end{figure}

In Figure \ref{fig:RetrievalAccu_Noise_Temp_Dyn}, below, we keep the system's parameters as above and we choose: $\tau_1=\tau_2 = 0.6,\kappa=0.1, \lambda = 0.05,\Delta t=0.1$.
The results presented in this study highlight the potential of nonlinear plasticity models, particularly the Allee-based framework, to enhance memory capacity and robustness in neural networks. The Allee model introduces critical nonlinear feedback mechanisms that address several limitations of traditional models such as Hebbian and Oja’s rules. This is evident in its ability to stabilize weights, suppress unbounded growth, and while preserving its ability to  differentiate between stored patterns. Such features make it a promising approach for exploring neural dynamics, especially in noisy environments.
\section{Discussion}\label{sec:concl}
In this study, we developed and analyzed a nonlinear synaptic plasticity model grounded in the Allee effect, offering a compelling alternative to classical frameworks such as the Hebbian and Oja learning rules. Our exploration demonstrated that incorporating nonlinear feedback mechanisms and critical threshold dynamics can significantly enhance both the stability and robustness of memory storage and pattern retrieval in neural networks. By constructing and analyzing both a simplified single-neuron model and an extended multilayer network, we conducted a comprehensive theoretical and numerical investigation of the model’s behavior under a wide variety of conditions.

A central innovation of this work lies in the nonlinear regularization of synaptic weights via the Allee threshold, which suppresses unbounded weight growth and introduces meaningful extinction dynamics. Our bifurcation analysis revealed that the Allee model supports multiple regimes of behavior--including stable fixed points, extinction states, and oscillatory activity--depending on the parameter choices. In particular, the model admits saddle-node, transcritical, and Hopf bifurcations, giving it the capacity to capture a wide range of biological and computational phenomena, from bi-stability to rhythmic firing patterns.

Our theoretical results were complemented by detailed simulations that demonstrate how the Allee model enhances memory capacity and resilience to noise. The model’s ability to retain or forget patterns based on threshold-driven dynamics offers a biologically plausible mechanism for adaptive memory. Unlike the Hebbian and Oja rules, which are  unable to suppress weak synaptic weights, the Allee model selectively filters subthreshold inputs and promotes strong, sustained activations. This mechanism approximates memory consolidation processes observed in biological systems in an intuitive and effective manner.

We also introduced temporal extensions of the Allee model by integrating eligibility traces. These modifications  bring the model closer to biological realism by allowing past neural activity to influence synaptic updates over time. The resulting dynamics improve pattern retrieval accuracy, especially in noisy environments, and narrow the performance gap between the Allee model and spike-timing-dependent plasticity rules (STDP), which already benefit from embedded temporal sensitivity.

Moreover, we conducted a rigorous comparison between the Allee, Hebbian, Oja, and STDP  models in order to highlight  each framework’s strengths and limitations. The Allee model performs as well and in certain cases outperforms  the classical Hebbian and Oja models  in scenarios involving weight stabilization, and bi-stability. 
In sum, this work provides a comprehensive foundation for understanding how nonlinear synaptic rules inspired by ecological dynamics (like the Allee effect) can inform the design of more robust, stable, and biologically grounded learning systems.

The Allee-based plasticity model proposed in this paper is not intended as a direct replication of synaptic processes observed in neural tissue. Instead, it serves as a conceptual framework that leverages ecological dynamics--specifically threshold-based feedback mechanisms--to explore how memory and pattern retrieval might be stabilized under nonlinear constraints. This abstraction allows for mathematical tractability while offering biologically informed insights into weight regulation, robustness, and adaptive behavior in neural system.

Looking ahead, several avenues for future exploration emerge. First, empirical validation of the Allee plasticity rule in biological settings would enhance its scientific relevance and applicability. Second, expanding the model to include realistic spiking dynamics, inhibitory populations, and recurrent topologies could further align it with observed cortical behavior. Third, integrating the Allee framework into artificial intelligence systems--particularly for tasks involving continual learning or pattern completion under noise--may yield more adaptive and interpretable neural architectures.
Finally, this work opens the door to a unified theory of memory and plasticity that integrates nonlinear dynamical systems theory with biological realism. The Allee model, especially in its temporally extended form, provides a promising blueprint for neural systems that are both adaptive and selective, offering resilience in the face of uncertainty while preserving the capacity for complex learning.
\bibliography{AlleeMemory}
\section*{Appendix}
\subsection*{Proof of Theorem \ref{thm1}} \label{App:A}

\[
J(x, y) =
\begin{bmatrix}
\frac{\partial f}{\partial x} & \frac{\partial f}{\partial y} \\
\frac{\partial g}{\partial x} & \frac{\partial g}{\partial y}
\end{bmatrix}.
\]
The partial derivatives of \(f(x, y)\) are:
\[
\frac{\partial f}{\partial x} = -1 + m G'(u\sqrt{y} + mx),
\]
\[
\frac{\partial f}{\partial y} = \frac{u}{2\sqrt{y}} G'(u\sqrt{y} + mx).
\]
The partial derivatives of \(g(x, y)\) are:
\[
\frac{\partial g}{\partial x} = \left(u\sqrt{y} - \frac{2xy}{K}\right)\left(1 - \frac{A}{y}\right),
\]
\[
\frac{\partial g}{\partial y} = x\left(\frac{u}{2\sqrt{y}}-\frac{x}{K}\right) \left(1 - \frac{A}{y}\right) + x\left(u\sqrt{y} - \frac{xy}{K}\right)\left(\frac{A}{y^2}\right).
\]
%\]
The eigenvalues of the Jacobian matrix are:
\[
\lambda_{1,2} = \frac{\text{tr}(J) \pm \sqrt{(\text{tr}(J))^2 - 4 \det(J)}}{2},
\]
where:
\begin{eqnarray*}
tr(J):=tr(x,y) &=&\frac{\partial f}{\partial x}+ \frac{\partial g}{\partial y}\\
&=&  -1 + m G'(u \sqrt{y} + m x) +\left(\frac{ux}{2\sqrt{y}}-\frac{x^2}{K}\right)\left( 1-\frac{A}{y}\right)+\left(ux\sqrt{y}-\frac{x^2y}{K}\right)\frac{A}{y^2}\;.
\end{eqnarray*}

and 
\begin{eqnarray*}
det(J):=det(x,y)&=& \frac{\partial f}{\partial x} \frac{\partial g}{\partial y} - \frac{\partial f}{\partial y} \frac{\partial g}{\partial x}\\
&=& \left[-1 + m G'(u \sqrt{y} + m x)\right]\left[\left(\frac{ux}{2\sqrt{y}}-\frac{x^2}{K}\right)\left( 1-\frac{A}{y}\right)+\left(ux\sqrt{y}-\frac{x^2y}{K}\right)\frac{A}{y^2}\right]\\
&-& \frac{u}{2\sqrt{y}}G'(u \sqrt{y} + m x)
\left(u\sqrt{y} - \frac{2 x y}{K}\right) \left(1 - \frac{A}{y}\right)\;.\\
\end{eqnarray*}
\noindent For $A\neq0$, there are two interior  fixed points. Therefore, we will only discuss the local stability of $(x^*=G(v_A), y^*=A)$. 
 Put \[v_A=u\sqrt{A}+mx_A, \quad \tau_A=\frac{uK}{\sqrt{A}}\;.\] When  \(y = A\), there exists $0<x_A<1$ such that $x_A=G(v_A)$, since $0<G(v)<1$ for all $v\in \R$.   In this case, we have that 
\[
\frac{\partial g}{\partial x} = 0, \quad \frac{\partial g}{\partial y} = x_A\left(\frac{u}{\sqrt{A}}- \frac{x_A}{K}\right).
\]
It follows that 
\[ \det(J) = \frac{\partial f}{\partial x}\frac{\partial g}{\partial y}\;.
\]
Hence, we have 
\[
\Delta=(\text{tr}(J))^2 - 4 \det(J)=\left(\frac{\partial f}{\partial x}+\frac{\partial g}{\partial y}\right)^2-4\frac{\partial f}{\partial x}\frac{\partial g}{\partial y}=\left(\frac{\partial f}{\partial x}-\frac{\partial g}{\partial y}\right)^2\;.
\] 
Thus the eigenvalues are given as 
\[
\lambda_{1,2} = \frac{ \frac{\partial f}{\partial x}+\frac{\partial g}{\partial y}\pm \abs{\frac{\partial f}{\partial x}-\frac{\partial g}{\partial y}}}{2}\;.
\]
Clearly, if $\ds  \frac{\partial f}{\partial x}>\frac{\partial g}{\partial y}$, then at $(x_A,A)$, we have  \[\ds \lambda_{1}= \frac{\partial f}{\partial x}=-1 + m G'(v_A),\quad   \lambda_2=\frac{\partial g}{\partial y}= x_A\left(\frac{u}{\sqrt{A}} - \frac{x_A}{K}\right).\] 
If $\ds  \frac{\partial f}{\partial x}<\frac{\partial g}{\partial y}$, then \[\ds \lambda_{1}= \frac{\partial g}{\partial y}=x_A\left(\frac{u}{\sqrt{A}} - \frac{x_A}{K}\right),\quad   \lambda_2=\frac{\partial f}{\partial x}=-1 + m G'(v_A)\;.\] 
The two cases are therefore interchangeable. On one hand,  $\lambda_1=x_A\left(\frac{u}{\sqrt{A}} - \frac{x_A}{K}\right)<0$ if $x_A\notin \left(0,\tau_A\right)$ and $\lambda_1>0$ if $x_A \in \left(0,\tau_A\right)$. We note that since we require that $0<x_A<1$, the condition $x_A\notin \left(0,\tau_A\right)$ is equivalent to $x_A\in \left(\tau_A,1\right)$. On the other hand, the sigmoid function \(G(v)\) has  derivative \(G'(v) = G(v)(1 - G(v))\). We know that since  $0<G(v)<1$ and that $G'(v)=G(v)(1 - G(v))$  has maximum $\ds \frac{1}{4}$.   It  follows that 
\[
0\leq G'(v)=G(v)(1-G(v))\leq \frac{1}{4}\;.
\]
Hence 
\[
-1\leq \lambda_2=-1 + m G'(v_A)\leq -1+\frac{m}{4}\;.
\]
Consequently, if $0<m<4$, then $\lambda_2<0$. 

\noindent To find conditions under which $\lambda_2>0$, we discuss the case when $ G'(v_A)>\frac{1}{m}$. \\
We note that $G'(v_A)$ is a second order polynomial in $G(v_A)$ defined on the interval $[0,1]$. 
\begin{figure}[H] 
   \centering
   \includegraphics[scale=0.5]{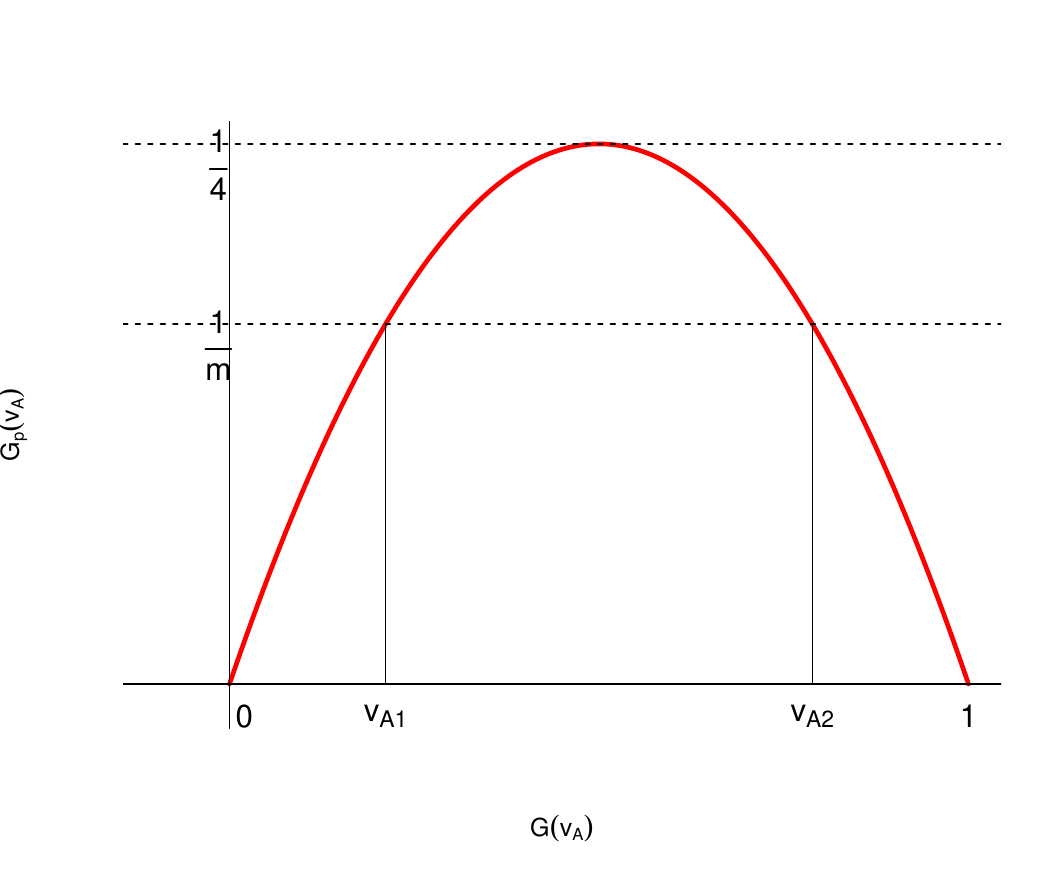} 
   \caption{Graph of $G'(v_A)$ as a function of $G(v_A)$.}
   \label{fig:sigmoidPlot}
\end{figure}
Therefore, if $m>4$, then $\ds \frac{1}{m}<\frac{1}{4}$. Since $\ds G'(v_A)\leq \frac{1}{4}$.  From Figure  \ref{fig:sigmoidPlot} above,   it follows that the horizontal line $y=\frac{1}{m}$ intersects the graph of $G'(v_A)$ at two points: $0<v_{A1}<v_{A2}<1$. Since the leading coefficient of the second order polynomial $G'(v_A)$ is negative, we infer that $G'(v_A)>\frac{1}{m}$ if $G(v_A)\in (v_{A1}, v_{A2})$ and $G'(v_A)<\frac{1}{m}$  if $G(v_A)\in (0, v_{A1})\cup (v_{A2}, 1)$. 
recalling that $x_A=G(v_A)$, we conclude that the interior fixed point $(x_A, A)$ is locally asymptotically stable ( $\lambda_1, \lambda_2<0$) and the other interior  fixed point   $(x_1^*, y_1^*)$ is unstable
 \begin{itemize}
 \item if $m<4$ and $\tau_A<x_A<1$,  or 
 \item if $m>4$ and $x_A \in (0, v_{A1})\cup (v_{A1}, 1)$. 
 \end{itemize}
It is unstable otherwise. 
%%%%%%%%%%%%%%%%%%%%%%%%%%%%%%%%%%%%%%%%%%%%%%%%%%%%%%%%%%%%%%%%%%%
\subsection*{Proof of Theorem \ref{thm2}} \label{App:B}
We have that 
     \begin{eqnarray*}
tr(x, y) &=&\frac{\partial f}{\partial x}+ \frac{\partial g}{\partial y}\;.
\end{eqnarray*}
And 
\begin{eqnarray*}
det(x,y) &=& \frac{\partial f}{\partial x} \frac{\partial g}{\partial y} - \frac{\partial f}{\partial y} \frac{\partial g}{\partial x}\;.
\end{eqnarray*}
      \subsubsection*{Case 1: $y^*=A$}
In this case $\ds \frac{\partial g}{\partial x}=0$ and therefore $\ds det(x^*,A)=\frac{\partial f}{\partial x} \frac{\partial g}{\partial y}$. It follows that if $tr(x^*,A)=0$, then $\ds \frac{\partial f}{\partial x} =-\frac{\partial g}{\partial y}$ and therefore, \[\det(x^*,A)=-\left(\frac{\partial f}{\partial x}\right)^2=-\left( \frac{\partial g}{\partial y}\right)^2<0\;.\]
Therefore, a Hopf bifurcation cannot occur in this case. 
      \subsubsection*{Case 2: $y^*\neq A$ and  $y^*=\left(\frac{uK}{x^*}\right)^2$}
    This  corresponds to $\ds u\sqrt{y^*}=\frac{x^*y^*}{K}$ or equivalently to $\ds \frac{u}{\sqrt{y}}=\frac{x^*}{K}$ and $uK=x^*\sqrt{y^*}$
. \\Put $v^*=u\sqrt{y^*}+mx^*$       Then it can be shown that 
      \begin{eqnarray*}
      tr(x^*, y^*)&=& -1 + m G'(v^*) +\left(\frac{ux^*}{2\sqrt{y^*}}-\frac{(x^*)^2}{K}\right)\left( 1-\frac{A}{y^*}\right)+\left(ux^*\sqrt{y}-\frac{(x^*)^2y^*}{K}\right)\frac{A}{(y^*)^2}\\
      &=& -1 + m G'(v^*) +\left(\frac{ux^*}{2\sqrt{y^*}}-\frac{(x^*)^2}{K}\right)\left( 1-\frac{A}{y^*}\right)\\
      &=& -1 + m G'(v^*) +x^*\left(\frac{u}{2\sqrt{y^*}}-\frac{(x^*}{K}\right)\left( 1-\frac{A}{y^*}\right)\\
        &=& -1 + m G'(v^*)-x^*\left( \frac{x^*}{2K}-\frac{x^*}{K}\right)\left( 1-\frac{A}{y^*}\right)\\
          &=& -1 + m G'(v^*)-\frac{(x^*)^2}{2K}\left( 1-\frac{A}{y^*}\right)\;.\\
      \end{eqnarray*}
      We also have that 
 \begin{eqnarray*}
det(x^*,y^*) &=& \left[-1 + m G'(v^*)\right]\left[x^*\left(\frac{u}{2\sqrt{y^*}}-\frac{x^*}{K}\right)\left( 1-\frac{A}{y^*}\right)+x^*\left(u\sqrt{y^*}-\frac{x^*y^*}{K}\right)\frac{A}{(y^*)^2}\right]\\
&-& \frac{u}{2\sqrt{y^*}}G'(v^*)\left(u\sqrt{y^*} - \frac{2 x^* y^*}{K}\right) \left(1 - \frac{A}{y^*}\right)\\\
&=&  \left[-1 + m G'(v^*)\right]\left[-\frac{(x^*)^2}{2K}\left( 1-\frac{A}{y^*}\right)\right]
- \frac{u}{2\sqrt{y^*}}G'(v^*)\left(- \frac{x^* y^*}{2K}\right) \left(1 - \frac{A}{y^*}\right)\\
&=& \frac{1}{2K}\left(1 - \frac{A}{y^*}\right)\left[-(x^*)^2(-1+mG'(v^*))+ux^*\sqrt{y^*}G'(v^*)\right]\\
&=& \frac{1}{2K}\left(1 - \frac{A}{y^*}\right)\left[G'(v^*)(-mx^2+u^2K)+x^2\right]\\
&=& \frac{1}{2K}\left(1-\frac{A}{\ys}\right)\left[(\xs)^2-m(\xs)^2G'(v^*)+u^2KG'(v^*)\right]\;.
\end{eqnarray*}
Therefore $\ds tr(x^*, y^*)=0\implies mG'(v^*)=1+\frac{(x^*)^2}{2K}\left(1 - \frac{A}{y^*}\right)$. Thus 
\[\ds m(x^*)^2G'(v^*)=(x^*)^2+\frac{(x^*)^4}{2K}\left(1 - \frac{A}{y^*}\right)\] and 
\[\ds u^2KG'(v^*)=\frac{u^2K}{m}\left[1+\frac{(x^*)^2}{2K}\left(1 - \frac{A}{y^*}\right)\right]\;.\]
In this case, the determinant simplifies to 
\begin{eqnarray*}
det(\xs,\ys)&=& \frac{1}{2K}\left(1-\frac{A}{\ys}\right)\left[-\frac{1}{2K}\left(1-\frac{A}{\ys}\right)(\xs)^4+\frac{u^2K}{2m}\left(1-\frac{A}{\ys}\right)(\xs)^2+\frac{u^2K}{m}\right]\;.
\end{eqnarray*}
Let  $\ds \lambda=1-\frac{A}{\ys}, \beta=\frac{u^2K^2}{2m}$, and $p=x^2$. Then 
\[det(\xs,\ys)=\frac{1}{4K^2}\lambda[- \lambda p^2+\beta \lambda p+2\beta]\;.\]
So  $det(\xs,\ys)>0$ if $\lambda<0$ and $- \lambda p^2+\beta \lambda p+2\beta<0$ or $\lambda>0$ and $- \lambda p^2+\beta \lambda p+2\beta>0$.
\subsubsection*{Case 2.1 $y^*<A$}
This corresponds to $\lambda<0$. Therefore, $- \lambda p^2+\beta \lambda p+2\beta<0$ if its discriminant  $\Delta>0$ and $p\in (p_1,p_2)$, where $p_i, i=1,2$ is a root of $det(\xs,\ys)$. We have that \[\Delta=(\lambda \beta)^2+8\lambda \beta=\beta\lambda (\beta \lambda+8)\;.\]
$\Delta>0$ if and only if  $\beta \lambda+8<0$. Then the roots of the polynomials are 
\[p_1=\frac{\beta}{2}-\sqrt{\frac{\beta^2}{4}+\frac{2\beta}{\lambda}}, \quad p_2=\frac{\beta}{2}+\sqrt{\frac{\beta^2}{4}+\frac{2\beta}{\lambda}}\;. \]
$p_1<0$, therefore the valid  interval for $p$ is $(0,p_2)$. In terms of $x^*$, we have that $0<(x^*)^2<p_2$. Since $\ys<A$ is equivalent to $\frac{(uK)^2}{x^2}<A$ or $\ds \frac{(uK)^2}{A}<(\xs)^2$. Combining both conditions, we obtain
 \[\frac{uK}{\sqrt{A}}<\xs<\sqrt{p_2}\;.\]
 
 \subsubsection*{Case 2.2 $y^*>A$}
 This corresponds to $\lambda>0$. Therefore, $- \lambda p^2+\beta \lambda p+2\beta>0$ if its discriminant  $\Delta<0$  or $\Delta>0$ and $p\in (-\infty, p_1)\cup(p_2,\infty)$. \\
$\Delta<0$ if and only $\beta \lambda+8<0$, that is,  $\ds \lambda<\frac{-8}{\beta}$. Since $\beta>0$, this implies $\lambda<0$, which contradicts the assumption that $\lambda>0$.\\

 $\Delta>0$ if and only $\beta \lambda+8>0$, with roots $p_1, p_2$. 
 Since $p_1<0$ and $p_2>0$, the valid interval for $p$ is $(p_2,\infty)$. In terms of $x^*$, this corresponds to $x^*>\sqrt{p_2}$.

\end{document}